\documentclass[a4paper,fleqn]{cas-sc}

\usepackage[numbers]{natbib}
\usepackage{amsmath,amssymb,bm}
\usepackage{amsthm}
\usepackage{algorithm}
\usepackage{algorithmic}
\usepackage{booktabs}
\usepackage{float}
\usepackage{xcolor}
\usepackage[section]{placeins}
\usepackage{flafter}
\usepackage{capt-of}

\theoremstyle{remark}

\newcommand{\safeincludegraphics}[2][]{%
  \IfFileExists{#2}{\includegraphics[#1]{#2}}{%
    \fbox{\parbox[c][0.25\textheight][c]{0.9\linewidth}{\centering Missing figure: \texttt{#2}}}%
  }%
}
\newenvironment{ack}{\section*{Acknowledgements}}{}

\def\tsc#1{\csdef{#1}{\textsc{\lowercase{#1}}\xspace}}
\tsc{WGM}
\tsc{QE}
\tsc{EP}
\tsc{PMS}
\tsc{BEC}
\tsc{DE}
\begin{document}
\let\WriteBookmarks\relax
\def\floatpagepagefraction{1}
\def\textpagefraction{.001}
\shorttitle{Online identification of nonlinear time-varying systems}
\shortauthors{H. Ren et~al.}
%\begin{frontmatter}

\title [mode = title]{Online sparse Bayesian identification of nonlinear time-varying systems}                      
\tnotemark[1]

\tnotetext[1]{This paper was not presented at any conference.}

\author[tyut]{He Ren}
\ead{renhe0068@link.tyut.edu.cn}
\credit{Conceptualization, Methodology, Software, Validation, Writing -- original draft}

\author[tyut]{Gaowei Yan}
\cormark[1]
\ead{yangaowei@tyut.edu.cn}
\credit{Conceptualization, Supervision, Funding acquisition, Writing -- review and editing}

\author[tyut]{Hang Liu}
\ead{2023310084@link.tyut.edu.cn}
\credit{Software, Data curation, Validation}

\author[tyut]{Lifeng Cao}
\ead{caolifeng0087@link.tyut.edu.cn}
\credit{Investigation, Formal analysis, Validation}

\author[tyut]{Gang Dang}
\ead{danggang@tyut.edu.cn}
\credit{Supervision, Project administration}

\author[gemeng]{Zhujun Zhu}
\ead{zzhu_jun@163.com}
\credit{Resources, Project administration, Validation}

\affiliation[tyut]{organization={College of Electrical and Power Engineering, Taiyuan University of Technology},
                   city={Taiyuan},
                   postcode={030024},
                   country={China}}

\affiliation[gemeng]{organization={Shanxi Gemeng China-U.S. Clean Energy Research and Development Center Co., Ltd.},
                      city={Taiyuan},
                      postcode={030024},
                      country={China}}

\cortext[cor1]{Corresponding author.}

\begin{abstract}
Sparse regression provides a compact and interpretable route for nonlinear system modeling by selecting a small number of active terms from a candidate dictionary. Most sparse regressors, however, are constructed offline and then used as static predictors. In online operation, changing load, material properties, ambient conditions, or equipment states may alter both the coefficient values and the effective active support within the dictionary. Moreover, a direct recursive update over a rich dictionary may spread the adaptation over many weakly relevant terms, causing an initially sparse model to become increasingly dense. The key problem is therefore to maintain a sparse regressor online, so that it can absorb streaming data while keeping a compact but revisable active structure. This paper develops a Bayesian recursive sparse learning (BRSL) method for online sparse identification over candidate dictionary terms. The coefficient distribution is updated through a Bayesian posterior recursion, where sliding-window likelihood-ratio information recursion incorporates new samples, removes expired samples, and discounts historical information in a unified update. To preserve sparsity during recursion, posterior-guided shrinkage is introduced to suppress weakly supported dictionary terms and revise the active structure according to posterior evidence. The posterior update is performed in a candidate subspace with an adaptive information floor to keep the recursive solve well posed, and a bounded-error relation is given to clarify the influence of shrinkage, residual information, coefficient drift, and information conditioning. The proposed method is evaluated on sparse coefficient tracking and a power-plant-oriented multi-input multi-output (MIMO) nonlinear time-varying identification benchmark. The results show that the method maintains accurate online prediction, compact active supports, low off-support coefficient leakage, and informative posterior uncertainty under changing operating conditions.
\end{abstract}

% \includegraphics{figs/cas-grabs.pdf}

% \begin{highlights}
% \item A Bayesian recursive sparse learning method is developed for online nonlinear system identification.
% \item Sliding-window likelihood-ratio information recursion updates posterior information from streaming data.
% \item Posterior-guided shrinkage maintains compact active supports under coefficient drift.
% \item Candidate-subspace posterior updates avoid large Kronecker information matrices for MIMO systems.
% \item Numerical studies show accurate prediction, low coefficient leakage, and informative uncertainty.
% \end{highlights}

\begin{keywords}
Nonlinear time-varying systems \sep Online sparse Bayesian identification \sep Group horseshoe prior \sep Uncertainty quantification
\end{keywords}

\maketitle

\section{Introduction}
\label{sec:introduction}

In process and energy systems, nonlinear dynamic models are widely used to support soft sensing, monitoring, optimization, and control \citep{DOCHAIN2003801,Khatibisepehr2013,QIN2003733}. 
In these applications, the model is not only expected to fit historical data, but also to remain reliable when it is used online with streaming measurements. 
This requirement is difficult to satisfy because the data distribution encountered after deployment may differ from that used during model construction. 
Changes in load demand, feed properties, ambient conditions, fouling, equipment aging, or sensor drift can alter the relation between measured variables and system dynamics \citep{Kadlec2009,Ge2013,LU201823}. 
For a nonlinear model built from a rich candidate dictionary, such changes may affect both the coefficient values and the effective active terms needed to describe the current dynamics. 
Meanwhile, recursive adaptation without sparsity maintenance may distribute the update over many weakly relevant terms and gradually turn an initially compact model into a dense and opaque predictor. 
The difficulty is therefore to adapt the model online while preserving a sparse structure that remains useful for interpretation, maintenance, and controller-oriented use.

Sparse regression provides a natural structural basis for this problem. 
It represents nonlinear dynamics by selecting a small number of active terms from a candidate dictionary, where the dictionary may be constructed from physical relations, empirical nonlinearities, interaction terms, or other basis functions. 
Classical sparse regularization laid the foundation for sparse model selection \citep{Tibshirani1996}, and sparse identification of nonlinear dynamics showed that compact governing equations can be recovered from rich candidate libraries \citep{Brunton2016,kaiser2017sparse}. 
Sparse kernel-based regularization has also been used for parsimonious system identification \citep{Chen2014}, and recent studies summarize and extend sparse regularization in system identification and related dynamic modeling problems \citep{pillonetto2022sparse,liao2025data}. 
Bayesian sparse learning further combines coefficient shrinkage with uncertainty quantification \citep{ParkCasella2008,Carvalho2009,Dai2023}, while horseshoe-type priors provide strong shrinkage for small coefficients and reduced bias for large signals \citep{Bhadra2019}. 
These studies make sparse regression an effective framework for building compact and interpretable nonlinear models from candidate dictionaries.

Most sparse regression models, however, are constructed offline. 
The active terms and their coefficients are identified from a fixed data set and then used as a static predictor. 
This static use becomes restrictive when the operating condition moves away from the calibration data. 
Model degradation and maintenance have long been discussed in data-driven soft sensing and process monitoring \citep{Kadlec2009,Ge2013}. 
Bayesian methods have also been reviewed for inferential sensing in process applications, where uncertainty and reliability are important for industrial deployment \citep{Khatibisepehr2013}. 
Industrial online soft-sensor studies further show that model validity can deteriorate when operating regimes deviate from the calibration data \citep{Deng2013,LU201823}. 
In this situation, the sparse dictionary may still provide a meaningful representation, but the offline sparse model is no longer necessarily adequate because its selected terms and coefficients may become outdated. 
The selected coefficients may drift, and some previously weak or inactive terms may become relevant under the new operating condition. 
This creates a need to maintain both coefficient values and a compact active structure during online operation.

Recursive estimation provides a natural way to adapt models from streaming data. 
Kalman filtering provides the classical recursive framework for updating estimates as new observations arrive \citep{Kalman1960}, and unscented filtering extends this idea to nonlinear estimation problems \citep{JulierUhlmann2004}. 
Sequential Monte Carlo methods further provide a general recursive estimation route for nonlinear and non-Gaussian state-space problems \citep{Kantas2015}. 
Recent recursive-estimation studies continue to develop Kalman-filter variants, likelihood-based recursive estimators, and security-aware recursive state estimation for modern dynamic systems \citep{yang2025fractional,surya2024maximum,zou2025recursive}. 
In system identification, recursive least-squares-type algorithms have been developed for nonlinear model classes such as Wiener systems \citep{Ding2016}. 
Adaptive recursive identification has also been studied for time-varying Wiener output-error systems with unknown noise statistics \citep{Wang2020}. 
Related state and parameter estimation problems for bilinear state-space systems have been investigated in \citep{Zhang2018}, and multirate process identification with time delay has been treated using expectation--maximization methods \citep{Gu2019}. 
These recursive methods are effective for streaming adaptation, but they are usually applied to a prescribed model structure. 
When all terms in a rich nonlinear dictionary are updated recursively, the adaptation may be distributed over many weakly relevant terms, so that short-term prediction is improved at the cost of losing sparsity. 
Existing sparse-identification and digital-twin studies have begun to address time-varying models, state estimation with unknown governing equations, digital-twin generalization, multi-task and closed-loop settings, switching nonlinear systems, time-delay Hammerstein structures, and engineered sparse dictionaries \citep{wang2022time,wang2023simultaneous,edington2023time,course2023state,zhang2025caper,Zhang2023,zhang2025identification,yu2025structure,liao2025data}. 
However, many of these methods are still based on batch estimation, repeated sparse refitting, or recursive updates in which sparse support maintenance is not the central object. 
Repeated refitting may make the selected support sensitive to local excitation and correlated dictionary terms, while dense recursive updating may erode the compact structure of the offline sparse model. 
Thus, the key problem considered in this paper is to maintain an uncertainty-aware sparse regressor online, with drifting coefficients and a revisable effective active support.

This paper develops a Bayesian recursive sparse learning (BRSL) method for this online sparse-maintenance problem. 
The main update is formulated as a Bayesian posterior recursion for the coefficient distribution over candidate terms. 
To account for the influence of historical information under changing operating conditions, a sliding-window likelihood-ratio information recursion is derived for posterior updating. 
To avoid dense growth during recursive adaptation, posterior-guided shrinkage is introduced to suppress weakly supported dictionary terms and revise the active structure according to posterior evidence. 
The posterior update is performed in a candidate subspace with an adaptive information floor to keep the recursive solve well posed, and a bounded-error relation is provided to clarify the influence of shrinkage, residual information, coefficient drift, and information conditioning.

The main contributions are summarized as follows.
\begin{itemize}
    \item An online Bayesian sparse learning framework is proposed to maintain uncertainty-aware sparse regressors, where the coefficients and effective active support are allowed to evolve after offline model construction.

    \item A sliding-window likelihood-ratio information recursion is developed for recursive posterior updating. Combined with posterior-guided shrinkage, it adapts the coefficient posterior from streaming data while suppressing dense recursive growth and preserving the ability to revise the active structure.

    \item A candidate-subspace posterior update with an adaptive information floor is introduced to keep the recursive update well posed and computationally tractable. A bounded-error relation is further provided for drifting coefficients under bounded residual and information conditions.
\end{itemize}

The method is evaluated on sparse coefficient tracking and a power-plant-oriented MIMO nonlinear time-varying identification benchmark, focusing on prediction accuracy, support maintenance, off-support coefficient leakage, and posterior uncertainty.

The remainder of this paper is organized as follows. Section~\ref{sec2} formulates the online sparse identification problem. Section~\ref{sec3} presents the BRSL method and theoretical analysis. Section~\ref{sec4} reports the numerical studies. Section~\ref{sec5} concludes the paper.

\section{Problem Formulation}
\label{sec2}

\subsection{Streaming data and dictionary representation}

Consider a process system observed at discrete sampling instants \(t=1,2,\ldots\). Let \(\boldsymbol{y}_t\in\mathbb{R}^{n_y}\) denote the measured output vector, and let \(\boldsymbol{x}_t\in\mathbb{R}^{n_x}\) denote a regressor vector constructed from measured states, manipulated inputs, disturbance variables, and other operating variables. At time \(t\), the available streaming data are denoted by
\begin{equation}
\label{eq:data_stream}
\mathcal{D}_t
=
\left\{
(\boldsymbol{x}_i,\boldsymbol{y}_i)
\right\}_{i=1}^{t}.
\end{equation}
When a matrix form is needed, the sampled regressors and outputs are stacked as
\begin{equation}
\label{eq:data_matrices}
\boldsymbol{X}_{1:t}
=
\begin{bmatrix}
\boldsymbol{x}_{1}^{\top}\\
\boldsymbol{x}_{2}^{\top}\\
\vdots\\
\boldsymbol{x}_{t}^{\top}
\end{bmatrix}
\in \mathbb{R}^{t\times n_x},
\qquad
\boldsymbol{Y}_{1:t}
=
\begin{bmatrix}
\boldsymbol{y}_{1}^{\top}\\
\boldsymbol{y}_{2}^{\top}\\
\vdots\\
\boldsymbol{y}_{t}^{\top}
\end{bmatrix}
\in \mathbb{R}^{t\times n_y}.
\end{equation}
where \(\boldsymbol{X}_{1:t}\) is a regressor matrix rather than a pure state matrix, since each \(\boldsymbol{x}_i\) may include state measurements, inputs, disturbances, and operating-condition variables.

The nonlinear input--output relation is represented by a candidate dictionary. Let
\begin{equation}
\label{eq:raw_dictionary}
\boldsymbol{\varphi}(\boldsymbol{x}_t)
=
\left[
\varphi_1(\boldsymbol{x}_t),
\varphi_2(\boldsymbol{x}_t),
\ldots,
\varphi_p(\boldsymbol{x}_t)
\right]^{\top}
\in\mathbb{R}^{p}
\end{equation}
denote the raw candidate dictionary. The candidate terms may include polynomial functions, interaction terms, known physical relations, or empirical basis functions selected from engineering knowledge. Since different candidate terms may have different numerical magnitudes and physical units, the dictionary is normalized before sparse Bayesian updating.

Let \(\mathcal{D}_0=\{(\boldsymbol{x}_i^0,\boldsymbol{y}_i^0)\}_{i=1}^{N_0}\) be an initial calibration data set or a representative historical data set. For each candidate term, define
\begin{equation}
\label{eq:dictionary_mean}
m_j
=
\frac{1}{N_0}
\sum_{i=1}^{N_0}
\varphi_j(\boldsymbol{x}_i^0),
\end{equation}
and
\begin{equation}
\label{eq:dictionary_scale}
s_j
=
\max
\left\{
\left[
\frac{1}{N_0-1}
\sum_{i=1}^{N_0}
\left(
\varphi_j(\boldsymbol{x}_i^0)-m_j
\right)^2
\right]^{1/2},
\epsilon_{\mathrm{sc}}
\right\},
\end{equation}
where \(\epsilon_{\mathrm{sc}}>0\) is a small constant used to avoid division by a near-zero scale. The normalized candidate term is then given by
\begin{equation}
\label{eq:normalized_candidate}
\psi_j(\boldsymbol{x}_t)
=
\frac{
\varphi_j(\boldsymbol{x}_t)-m_j
}{
s_j
},
\qquad j=1,\ldots,p.
\end{equation}
The normalized dictionary is denoted by
\begin{equation}
\label{eq:normalized_dictionary}
\boldsymbol{\psi}(\boldsymbol{x}_t)
=
\left[
\psi_1(\boldsymbol{x}_t),
\psi_2(\boldsymbol{x}_t),
\ldots,
\psi_p(\boldsymbol{x}_t)
\right]^{\top}.
\end{equation}

The output vector is also normalized using the calibration data set. Let \(\boldsymbol{m}_y\in\mathbb{R}^{n_y}\) denote the output mean vector and let \(\boldsymbol{s}_y=[s_{y,1},\ldots,s_{y,n_y}]^{\top}\) denote the output scaling vector, where each \(s_{y,q}\) is the sample standard deviation of the \(q\)-th output over \(\mathcal{D}_0\), lower bounded by \(\epsilon_{\mathrm{sc}}\). Define the diagonal scaling matrix
\[
\boldsymbol{D}_y
=
\operatorname{diag}(s_{y,1},\ldots,s_{y,n_y}) .
\]
The normalized output is
\begin{equation}
\label{eq:output_normalization}
\bar{\boldsymbol{y}}_t
=
\boldsymbol{D}_y^{-1}
(
\boldsymbol{y}_t-\boldsymbol{m}_y
).
\end{equation}
The normalization constants \(m_j\), \(s_j\), \(\boldsymbol{m}_y\), \(\boldsymbol{s}_y\), and \(\boldsymbol{D}_y\) are obtained from the calibration stage and then used consistently during recursive identification. In this way, subsequent changes in the estimated coefficients reflect coefficient drift in a common scaled coordinate system.

The online identification is carried out in the normalized coordinate system. The normalized output is modeled as
\begin{equation}
\label{eq:normalized_fixed_dictionary_model}
\bar{\boldsymbol{y}}_t
=
(\boldsymbol{B}_t^{\star})^{\top}
\boldsymbol{\psi}(\boldsymbol{x}_t)
+
\boldsymbol{e}_t,
\end{equation}
where \(\boldsymbol{B}_t^{\star}\in\mathbb{R}^{p\times n_y}\) is the ideal coefficient matrix at time \(t\) in the normalized dictionary space, and \(\boldsymbol{e}_t\in\mathbb{R}^{n_y}\) collects measurement noise and unmodeled residuals in the normalized output space. The \(j\)-th row of \(\boldsymbol{B}_t^{\star}\) represents the contribution of the normalized candidate term \(\psi_j(\boldsymbol{x}_t)\) to all output channels.

If prior process knowledge is available, the dictionary can be partitioned as
\begin{equation}
\label{eq:dictionary_partition}
\boldsymbol{\psi}(\boldsymbol{x}_t)
=
\left[
\boldsymbol{\psi}_{\mathrm{emp}}^{\top}(\boldsymbol{x}_t),
\boldsymbol{\psi}_{\mathrm{phy}}^{\top}(\boldsymbol{x}_t)
\right]^{\top},
\end{equation}
where \(\boldsymbol{\psi}_{\mathrm{emp}}\) denotes empirical candidate functions and \(\boldsymbol{\psi}_{\mathrm{phy}}\) denotes physics-motivated terms. This partition describes the origin of the candidate terms, while all coefficients are estimated within the same coefficient matrix.

\subsection{Linear-in-the-parameters form and local data block}

For Bayesian updating, Eq.~\eqref{eq:normalized_fixed_dictionary_model} is rewritten in a vectorized linear-in-the-parameters form. Define
\begin{equation}
\label{eq:beta_vec}
\boldsymbol{\beta}_t^{\star}
=
\operatorname{vec}(\boldsymbol{B}_t^{\star})
\in \mathbb{R}^{p n_y},
\end{equation}
and
\begin{equation}
\label{eq:Phi_t}
\boldsymbol{\Phi}_t
=
\boldsymbol{I}_{n_y}
\otimes
\boldsymbol{\psi}^{\top}(\boldsymbol{x}_t)
\in
\mathbb{R}^{n_y\times p n_y},
\end{equation}
where \(\otimes\) denotes the Kronecker product. Then Eq.~\eqref{eq:normalized_fixed_dictionary_model} becomes
\begin{equation}
\label{eq:linear_parameter_model}
\bar{\boldsymbol{y}}_t
=
\boldsymbol{\Phi}_t
\boldsymbol{\beta}_t^{\star}
+
\boldsymbol{e}_t.
\end{equation}
Although the model is nonlinear in the measured variables through \(\boldsymbol{\psi}(\boldsymbol{x}_t)\), it is linear in the coefficient vector \(\boldsymbol{\beta}_t^{\star}\). This property is used in the recursive Bayesian update developed in the next section.

For a recent data block, let
\begin{equation}
\label{eq:local_index_set}
\mathcal{I}_t
=
\{\max(1,t-L+1),\ldots,t\}
\end{equation}
denote the index set with window length \(L\). Since the coefficients may drift over time, the samples in \(\mathcal{I}_t\) do not necessarily share exactly the same coefficient vector. Under the slow-drift setting, they are represented locally around the current coefficient vector \(\boldsymbol{\beta}_t^{\star}\) as
\begin{equation}
\label{eq:block_model}
\bar{\boldsymbol{y}}_{\mathcal{I}_t}^{\mathrm{vec}}
=
\boldsymbol{\Phi}_{\mathcal{I}_t}
\boldsymbol{\beta}_t^{\star}
+
\widetilde{\boldsymbol{e}}_{\mathcal{I}_t}^{\mathrm{vec}},
\end{equation}
where
\begin{equation}
\label{eq:block_definitions}
\bar{\boldsymbol{y}}_{\mathcal{I}_t}^{\mathrm{vec}}
=
\begin{bmatrix}
\bar{\boldsymbol{y}}_{i_1}\\
\bar{\boldsymbol{y}}_{i_2}\\
\vdots\\
\bar{\boldsymbol{y}}_{i_m}
\end{bmatrix},
\quad
\boldsymbol{\Phi}_{\mathcal{I}_t}
=
\begin{bmatrix}
\boldsymbol{\Phi}_{i_1}\\
\boldsymbol{\Phi}_{i_2}\\
\vdots\\
\boldsymbol{\Phi}_{i_m}
\end{bmatrix},
\quad
\widetilde{\boldsymbol{e}}_{\mathcal{I}_t}^{\mathrm{vec}}
=
\begin{bmatrix}
\widetilde{\boldsymbol{e}}_{i_1}\\
\widetilde{\boldsymbol{e}}_{i_2}\\
\vdots\\
\widetilde{\boldsymbol{e}}_{i_m}
\end{bmatrix},
\end{equation}
with \(\mathcal{I}_t=\{i_1,i_2,\ldots,i_m\}\). Specifically, for each \(i\in\mathcal{I}_t\),
\begin{equation}
\label{eq:equivalent_residual}
\widetilde{\boldsymbol{e}}_i
=
\boldsymbol{e}_i
+
\boldsymbol{\Phi}_i
\left(
\boldsymbol{\beta}_i^{\star}
-
\boldsymbol{\beta}_t^{\star}
\right).
\end{equation}
where \(\widetilde{\boldsymbol{e}}_i\) is an equivalent residual that combines the original residual \(\boldsymbol{e}_i\) and the local approximation error caused by using the current coefficient vector \(\boldsymbol{\beta}_t^{\star}\) to represent the sample at time \(i\). When the coefficient vector is constant within the data block, \(\widetilde{\boldsymbol{e}}_i=\boldsymbol{e}_i\). Therefore, Eq.~\eqref{eq:block_model} is a local vectorized representation around the current operating condition. In Section~\ref{sec3}, the same block data are also written in matrix form as \(\bar{\boldsymbol Y}_{\mathcal I_t}\in\mathbb R^{m\times n_y}\) when this notation is more convenient for covariance-weighted posterior derivations.

\subsection{Coefficient drift in a candidate representation}

The time variation considered in this paper is coefficient drift in the candidate-dictionary representation. The ideal coefficient vector is assumed to evolve according to
\begin{equation}
\label{eq:coefficient_drift}
\boldsymbol{\beta}^{\star}_t
=
\boldsymbol{\beta}^{\star}_{t-1}
+
\boldsymbol{\omega}_t,
\end{equation}
where \(\boldsymbol{\omega}_t\in\mathbb{R}^{pn_y}\) denotes the coefficient change between two consecutive sampling instants. The drift is assumed to be bounded:
\begin{equation}
\label{eq:bounded_drift}
\|\boldsymbol{\omega}_t\|_2
\leq
\delta,
\qquad
0\leq \delta < \infty .
\end{equation}

Eq.~\eqref{eq:coefficient_drift} treats process nonstationarity as coefficient variation within a candidate representation. This setting is appropriate for operating periods in which the relevant physical or empirical terms remain meaningful, but their influence changes with load variation, operating-condition shifts, equipment aging, fouling, catalyst deactivation, or sensor drift. The dictionary \(\boldsymbol{\psi}(\cdot)\) then acts as a stable set of modeling coordinates, and online identification amounts to recursively tracking a sparse coefficient vector and its posterior uncertainty as new data arrive.

\subsection{Effective sparse coefficient structure}

Since the candidate dictionary may contain redundant or weakly relevant terms, the coefficient matrix is assumed to be effectively sparse at each operating condition. For the multi-output model in Eq.~\eqref{eq:normalized_fixed_dictionary_model}, relevance is considered at the dictionary-term level. Let \((\boldsymbol{b}_{j,t}^{\star})^{\top}\) denote the \(j\)-th row of \(\boldsymbol{B}_t^{\star}\). For a small relevance threshold \(\varepsilon_s>0\), the effective active support is defined as
\begin{equation}
\label{eq:effective_active_support}
S_{t,\varepsilon_s}^{\star}
=
\left\{
j\in\{1,\ldots,p\}:
\|\boldsymbol{b}_{j,t}^{\star}\|_2>\varepsilon_s
\right\}.
\end{equation}
The effective sparsity condition is written as
\begin{equation}
\label{eq:effective_sparsity_assumption}
|S_{t,\varepsilon_s}^{\star}|
\leq
s,
\qquad
s\ll p .
\end{equation}
Terms with coefficient norms below \(\varepsilon_s\) are regarded as weakly relevant in the normalized dictionary space. Thus, the effective support describes the candidate terms that make a practically significant contribution under the current operating condition.

In the vectorized representation, each dictionary term corresponds to a group of coefficients across all output channels. Specifically, the coefficient group associated with \(\psi_j(\boldsymbol{x}_t)\) is
\begin{equation}
\label{eq:coefficient_group}
\mathcal{G}_j
=
\{j,\; j+p,\; j+2p,\ldots,\; j+(n_y-1)p\}.
\end{equation}
Maintaining a compact model at the dictionary-term level is therefore equivalent to identifying a small number of relevant coefficient groups from \(\{\mathcal{G}_j\}_{j=1}^{p}\). In the proposed Bayesian recursive update, the estimated active set is determined from the posterior magnitude and uncertainty of these coefficient groups.

\subsection{Noise model and posterior objective}

The residual term in Eq.~\eqref{eq:linear_parameter_model} is modeled as
\begin{equation}
\label{eq:noise_model}
\boldsymbol{e}_t
\sim
\mathcal{N}(\boldsymbol{0},\boldsymbol\Sigma_p),
\qquad
\boldsymbol\Sigma_p\succ0.
\end{equation}
where \(\boldsymbol\Sigma_p\in\mathbb R^{n_y\times n_y}\) denotes the residual covariance among output channels in the normalized output coordinate.

Given the normalized streaming data
\begin{equation}
\label{eq:normalized_streaming_data}
\bar{\mathcal{D}}_t
=
\left\{
(\boldsymbol{\psi}(\boldsymbol{x}_i),\bar{\boldsymbol{y}}_i)
\right\}_{i=1}^{t},
\end{equation}
the objective is to recursively update the coefficient posterior and the active dictionary support. Specifically, at each time \(t\), the algorithm aims to obtain an approximate posterior distribution
\begin{equation}
\label{eq:posterior_objective}
p(\boldsymbol{\beta}_t|\bar{\mathcal{D}}_t)
\approx
\mathcal{N}
(
\boldsymbol{\mu}_t,
\boldsymbol{\Xi}_t
),
\end{equation}
together with an estimated active support \(S_t\), where \(\boldsymbol{\mu}_t\) and \(\boldsymbol{\Xi}_t\) denote the posterior mean and covariance of the coefficient vector in the normalized dictionary space. In the next section, this Gaussian form is obtained from a conditional Gaussian representation of the sparse Bayesian model.

The desired online identification procedure should satisfy three requirements. First, it should adapt the coefficient estimates as new measurements arrive under bounded coefficient drift. Second, it should preserve effective sparsity so that the model remains compact and interpretable during recursive updating. Third, it should retain posterior uncertainty information for assessing the reliability of the updated coefficients and predictions. The problem is therefore to update
\begin{equation}
\label{eq:update_target}
(
\boldsymbol{\mu}_t,
\boldsymbol{\Xi}_t,
S_t
)
\quad
\text{from}
\quad
(
\boldsymbol{\mu}_{t-1},
\boldsymbol{\Xi}_{t-1},
S_{t-1}
)
\end{equation}
using the newly available streaming data.

\section{Methodology}
\label{sec3}

This section develops the proposed BRSL method for the dictionary-based model formulated in Section~\ref{sec2}. The objective is to maintain, rather than repeatedly reconstruct, a sparse nonlinear regressor as streaming data arrive. At each update, the method performs four operations. First, the information carried by the current sliding window is updated by adding newly arrived samples and removing expired samples. Second, a conditional Gaussian posterior is solved on a candidate dictionary subspace. Third, posterior group moments are used to refresh the shrinkage precision and determine the active support reported by the sparse model. Finally, residual information is used to update the candidate subspace for the next posterior solve.

Three sets should be distinguished throughout this section. The full dictionary \(\{1,\ldots,p\}\) contains all candidate terms considered in the posterior update. The candidate set \(\mathcal C_t\subseteq\{1,\ldots,p\}\) is the computational subspace used in the posterior update at time \(t\). The active support \(\mathcal S_t\subseteq\mathcal C_t\) is the sparse support reported after posterior relevance screening. Therefore, a term being included in \(\mathcal C_t\) does not mean that it is declared active; it only means that this term is allowed to participate in the current posterior update.

At the beginning of update \(t\), the available recursive quantities are
\[
\boldsymbol\Lambda_t,\quad
\mathcal C_t,\quad
\mathcal S_{t-1},\quad
\boldsymbol G_{t-1}^{L},\quad
\boldsymbol Z_{t-1}^{L},
\]
where \(\boldsymbol\Lambda_t\) is the shrinkage precision produced at the previous update, \(\mathcal C_t\) is the current candidate set, \(\mathcal S_{t-1}\) is the previous active support, and \(\boldsymbol G_{t-1}^{L}\), \(\boldsymbol Z_{t-1}^{L}\) are the stored sliding-window information matrices. After the posterior update, the algorithm produces the posterior mean and covariance, the current active support \(\mathcal S_t\), the next shrinkage precision \(\boldsymbol\Lambda_{t+1}\), and the next candidate set \(\mathcal C_{t+1}\).

\subsection{Conditional Bayesian sparse model}
\label{subsec:conditional_bayesian_model}

Over a local window \(\mathcal I_t=\{i_1,\ldots,i_m\}\), the normalized dictionary-based model can be written as
\begin{equation}
\bar{\boldsymbol Y}_{\mathcal I_t}
=
\boldsymbol\Psi_{\mathcal I_t}\boldsymbol B_t
+
\boldsymbol E_{\mathcal I_t},
\label{eq:method_block_regression}
\end{equation}
where
\[
\bar{\boldsymbol Y}_{\mathcal I_t}
=
\begin{bmatrix}
\bar{\boldsymbol y}_{i_1}^{\top}\\
\cdots\\
\bar{\boldsymbol y}_{i_m}^{\top}
\end{bmatrix}
\in\mathbb R^{m\times n_y},
\qquad
\boldsymbol\Psi_{\mathcal I_t}
=
\begin{bmatrix}
\boldsymbol\psi^{\top}(\boldsymbol x_{i_1})\\
\cdots\\
\boldsymbol\psi^{\top}(\boldsymbol x_{i_m})
\end{bmatrix}
\in\mathbb R^{m\times p}.
\]
The coefficient matrix \(\boldsymbol B_t\in\mathbb R^{p\times n_y}\) is allowed to drift with the operating condition. The \(j\)-th row of \(\boldsymbol B_t\), denoted by \(\boldsymbol b_{j,t}^{\top}\), contains the coefficients of the \(j\)-th dictionary term for all output channels. Thus, sparsity is imposed at the dictionary-term level rather than on individual scalar coefficients.

Let
\begin{equation}
\boldsymbol\beta_t
=
\operatorname{vec}(\boldsymbol B_t),
\qquad
\boldsymbol Y_t
=
\operatorname{vec}(\bar{\boldsymbol Y}_{\mathcal I_t}),
\qquad
\boldsymbol\Phi_t
=
\boldsymbol I_{n_y}\otimes \boldsymbol\Psi_{\mathcal I_t}.
\label{eq:method_vectorization}
\end{equation}
Then Eq.~\eqref{eq:method_block_regression} becomes
\begin{equation}
\boldsymbol Y_t
=
\boldsymbol\Phi_t\boldsymbol\beta_t
+
\boldsymbol\varepsilon_t.
\label{eq:method_vectorized_model}
\end{equation}
The residual covariance across output channels is denoted by \(\boldsymbol\Sigma_e\succ0\). The block residual is modeled as
\begin{equation}
\boldsymbol\varepsilon_t
\sim
\mathcal N
\left(
\boldsymbol 0,
\boldsymbol\Sigma_e\otimes \boldsymbol I_m
\right).
\label{eq:method_noise_model}
\end{equation}

A group horseshoe prior is used to shrink irrelevant dictionary terms while allowing relevant terms to escape excessive shrinkage. Conditionally on the local and global scale variables, the prior for each coefficient group is written as
\begin{equation}
\boldsymbol b_{j,t}
\mid
\tau_t,\lambda_{j,t}
\sim
\mathcal N
\left(
\boldsymbol 0,
\tau_t^2\lambda_{j,t}^2\boldsymbol I_{n_y}
\right),
\qquad
\lambda_{j,t}\sim C^+(0,1),
\qquad
\tau_t\sim C^+(0,\tau_0).
\label{eq:method_group_horseshoe}
\end{equation}
Define the shrinkage precision
\begin{equation}
\lambda^{\rm sh}_{j,t}
=
(\tau_t^2\lambda_{j,t}^2)^{-1},
\qquad
\boldsymbol\Lambda_t
=
\operatorname{diag}
\left(
\lambda^{\rm sh}_{1,t},
\ldots,
\lambda^{\rm sh}_{p,t}
\right).
\label{eq:method_shrinkage_precision}
\end{equation}
Given \(\boldsymbol\Lambda_t\), the group horseshoe prior has the conditional Gaussian form
\begin{equation}
\boldsymbol\beta_t
\mid
\boldsymbol\Lambda_t
\sim
\mathcal N
\left(
\boldsymbol 0,
\boldsymbol I_{n_y}\otimes \boldsymbol\Lambda_t^{-1}
\right).
\label{eq:method_conditional_prior}
\end{equation}

Combining Eqs.~\eqref{eq:method_vectorized_model}--\eqref{eq:method_conditional_prior}, the conditional posterior is Gaussian:
\begin{equation}
p(\boldsymbol\beta_t|\boldsymbol Y_t,\boldsymbol\Lambda_t)
=
\mathcal N
(
\boldsymbol\mu_t,\boldsymbol\Xi_t
).
\label{eq:method_conditional_posterior}
\end{equation}
Let
\begin{equation}
\boldsymbol G_t
=
\boldsymbol\Psi_{\mathcal I_t}^{\top}
\boldsymbol\Psi_{\mathcal I_t},
\qquad
\boldsymbol Z_t
=
\boldsymbol\Psi_{\mathcal I_t}^{\top}
\bar{\boldsymbol Y}_{\mathcal I_t}.
\label{eq:method_batch_statistics}
\end{equation}
The posterior precision and information vector are
\begin{equation}
\boldsymbol J_t
=
\boldsymbol\Xi_t^{-1}
=
\boldsymbol\Sigma_e^{-1}\otimes \boldsymbol G_t
+
\boldsymbol I_{n_y}\otimes \boldsymbol\Lambda_t,
\label{eq:method_posterior_precision}
\end{equation}
\begin{equation}
\boldsymbol h_t
=
\operatorname{vec}
\left(
\boldsymbol Z_t\boldsymbol\Sigma_e^{-1}
\right),
\qquad
\boldsymbol\mu_t
=
\boldsymbol J_t^{-1}\boldsymbol h_t.
\label{eq:method_posterior_mean}
\end{equation}
These expressions follow from completing the square in the product of the Gaussian likelihood and the conditional Gaussian prior. Specifically, the log posterior contains the quadratic term
\[
-\frac{1}{2}
\boldsymbol\beta_t^{\top}
\left(
\boldsymbol\Sigma_e^{-1}\otimes\boldsymbol G_t
+
\boldsymbol I_{n_y}\otimes\boldsymbol\Lambda_t
\right)
\boldsymbol\beta_t
\]
and the linear term
\[
\boldsymbol\beta_t^{\top}
\operatorname{vec}
\left(
\boldsymbol Z_t\boldsymbol\Sigma_e^{-1}
\right),
\]
which gives Eqs.~\eqref{eq:method_posterior_precision} and \eqref{eq:method_posterior_mean}.

\subsection{Sliding-window likelihood-ratio information recursion}
\label{subsec:likelihood_ratio_recursion}

Directly recomputing Eqs.~\eqref{eq:method_batch_statistics}--\eqref{eq:method_posterior_mean} from the full window at every update is unnecessary. 
The proposed method therefore updates the window information recursively. 
Let \(\mathcal I_t\) denote the sample-index set of the current sliding window. 
The newly added and expired sample sets are defined as
\begin{equation}
    \mathcal A_t
    =
    \mathcal I_t\setminus \mathcal I_{t-1},
    \qquad
    \mathcal R_t
    =
    \mathcal I_{t-1}\setminus \mathcal I_t .
    \label{eq:method_added_removed_sets}
\end{equation}
Here, \(\mathcal A_t\) contains the samples that enter the current window, and \(\mathcal R_t\) contains the samples that are removed from the previous window. 
The current window information is obtained by discounting the retained historical information, adding the contribution of \(\mathcal A_t\), and subtracting the retained contribution of \(\mathcal R_t\).

Let \(\ell_i(\boldsymbol B)\) denote the Gaussian likelihood contribution of sample \(i\):
\begin{equation}
    \ell_i(\boldsymbol B)
    \propto
    \exp
    \left[
    -\frac{1}{2}
    \left(
    \bar{\boldsymbol y}_i
    -
    \boldsymbol B^{\top}\boldsymbol\psi_i
    \right)^{\top}
    \boldsymbol\Sigma_e^{-1}
    \left(
    \bar{\boldsymbol y}_i
    -
    \boldsymbol B^{\top}\boldsymbol\psi_i
    \right)
    \right],
    \label{eq:method_sample_likelihood}
\end{equation}
where \(\boldsymbol\psi_i=\boldsymbol\psi(\boldsymbol x_i)\). 
With a forgetting factor \(0<\xi\le1\), the weighted likelihood carried by the current sliding window is
\begin{equation}
    \mathcal L_t^L(\boldsymbol B)
    =
    \prod_{i\in\mathcal I_t}
    \ell_i(\boldsymbol B)^{\xi^{t-i}} .
    \label{eq:method_weighted_likelihood}
\end{equation}
Since the retained part of the previous window is discounted by \(\xi\), the current weighted likelihood can be written as the likelihood-ratio update
\begin{equation}
    \mathcal L_t^L(\boldsymbol B)
    \propto
    \left[
    \mathcal L_{t-1}^{L}(\boldsymbol B)
    \right]^{\xi}
    \frac{
    \prod_{i\in\mathcal A_t}
    \ell_i(\boldsymbol B)^{\xi^{t-i}}
    }{
    \prod_{i\in\mathcal R_t}
    \ell_i(\boldsymbol B)^{\xi^{t-i}}
    } .
    \label{eq:method_likelihood_ratio}
\end{equation}
Equation~\eqref{eq:method_likelihood_ratio} shows how the sliding-window likelihood is renewed: new likelihood contributions are incorporated, expired contributions are removed, and retained historical information is discounted.

For the Gaussian likelihood in Eq.~\eqref{eq:method_sample_likelihood}, the weighted likelihood is fully characterized by the sufficient statistics
\begin{equation}
\boldsymbol G_t^{L}
=
\sum_{i\in\mathcal I_t}
\xi^{t-i}
\boldsymbol\psi_i\boldsymbol\psi_i^{\top},
\qquad
\boldsymbol Z_t^{L}
=
\sum_{i\in\mathcal I_t}
\xi^{t-i}
\boldsymbol\psi_i\bar{\boldsymbol y}_i^{\top}.
\label{eq:method_weighted_statistics}
\end{equation}
The contributions of the newly added samples are
\begin{equation}
\boldsymbol G_{\mathcal A_t}
=
\sum_{i\in\mathcal A_t}
\xi^{t-i}
\boldsymbol\psi_i\boldsymbol\psi_i^{\top},
\qquad
\boldsymbol Z_{\mathcal A_t}
=
\sum_{i\in\mathcal A_t}
\xi^{t-i}
\boldsymbol\psi_i\bar{\boldsymbol y}_i^{\top},
\label{eq:method_added_statistics}
\end{equation}
and the retained contributions of the expired samples are
\begin{equation}
\boldsymbol G_{\mathcal R_t}
=
\sum_{i\in\mathcal R_t}
\xi^{t-i}
\boldsymbol\psi_i\boldsymbol\psi_i^{\top},
\qquad
\boldsymbol Z_{\mathcal R_t}
=
\sum_{i\in\mathcal R_t}
\xi^{t-i}
\boldsymbol\psi_i\bar{\boldsymbol y}_i^{\top}.
\label{eq:method_removed_statistics}
\end{equation}
Therefore, the information statistics are updated as
\begin{equation}
\boldsymbol G_t^{L}
=
\xi\boldsymbol G_{t-1}^{L}
+
\boldsymbol G_{\mathcal A_t}
-
\boldsymbol G_{\mathcal R_t},
\label{eq:method_G_recursion}
\end{equation}
\begin{equation}
\boldsymbol Z_t^{L}
=
\xi\boldsymbol Z_{t-1}^{L}
+
\boldsymbol Z_{\mathcal A_t}
-
\boldsymbol Z_{\mathcal R_t}.
\label{eq:method_Z_recursion}
\end{equation}
Equations~\eqref{eq:method_G_recursion} and \eqref{eq:method_Z_recursion} are the information-form recursion induced by the sliding-window likelihood-ratio update in Eq.~\eqref{eq:method_likelihood_ratio}.

For a fixed-length window 
\(\mathcal I_t=\{\max(1,t-L+1),\ldots,t\}\), one sample enters the window at each update, and one sample expires after the window is full. 
In this common case, Eqs.~\eqref{eq:method_G_recursion} and \eqref{eq:method_Z_recursion} reduce to
\begin{equation}
\boldsymbol G_t^{L}
=
\xi\boldsymbol G_{t-1}^{L}
+
\boldsymbol\psi_t\boldsymbol\psi_t^{\top}
-
\mathbf 1_{\{t-L\ge1\}}
\xi^L
\boldsymbol\psi_{t-L}\boldsymbol\psi_{t-L}^{\top},
\label{eq:method_G_single}
\end{equation}
\begin{equation}
\boldsymbol Z_t^{L}
=
\xi\boldsymbol Z_{t-1}^{L}
+
\boldsymbol\psi_t\bar{\boldsymbol y}_t^{\top}
-
\mathbf 1_{\{t-L\ge1\}}
\xi^L
\boldsymbol\psi_{t-L}\bar{\boldsymbol y}_{t-L}^{\top}.
\label{eq:method_Z_single}
\end{equation}
The factor \(\xi^L\) appears because the expired sample has remained in the window for \(L\) update steps.

Using the updated sufficient statistics, the raw full-dictionary posterior information before candidate restriction is
\begin{equation}
\boldsymbol J_t^{L,0}
=
\boldsymbol\Sigma_e^{-1}\otimes \boldsymbol G_t^{L}
+
\boldsymbol I_{n_y}\otimes \boldsymbol\Lambda_t,
\label{eq:method_raw_full_J}
\end{equation}
\begin{equation}
\boldsymbol h_t^{L,0}
=
\operatorname{vec}
\left(
\boldsymbol Z_t^{L}\boldsymbol\Sigma_e^{-1}
\right).
\label{eq:method_raw_full_h}
\end{equation}
The superscript \(0\) indicates that no positive-definiteness correction has yet been applied.

\subsection{Candidate-subspace posterior update}
\label{subsec:candidate_subspace_update}

The full dictionary may contain many weak or redundant terms, and solving the posterior over all \(p\) terms at every update is unnecessary. The posterior update is therefore performed on a candidate set \(\mathcal C_t\), which contains the previous active support and additional terms that may become relevant under the current operating condition. The candidate set is only a computational subspace. It should be distinguished from the active support \(\mathcal S_t\), which is the sparse structure reported after posterior relevance screening.

Let \(\boldsymbol P_{\mathcal C_t}\in\mathbb R^{p\times |\mathcal C_t|}\) be the column selection matrix that extracts the terms in \(\mathcal C_t\). In the vectorized multi-output representation, the corresponding selection matrix is
\begin{equation}
\boldsymbol T_{\mathcal C_t}
=
\boldsymbol I_{n_y}\otimes \boldsymbol P_{\mathcal C_t}.
\label{eq:method_candidate_selection}
\end{equation}
The raw candidate-subspace information matrix and information vector are obtained by projection:
\begin{equation}
\boldsymbol J_{\mathcal C,t}^{0}
=
\boldsymbol T_{\mathcal C_t}^{\top}
\boldsymbol J_t^{L,0}
\boldsymbol T_{\mathcal C_t},
\qquad
\boldsymbol h_{\mathcal C,t}^{0}
=
\boldsymbol T_{\mathcal C_t}^{\top}
\boldsymbol h_t^{L,0},
\label{eq:method_candidate_information}
\end{equation}
where the superscript \(0\) indicates that no numerical correction has yet been applied.

Because expired samples are explicitly discounted in the sliding-window recursion, the raw information matrix may become ill-conditioned on the candidate subspace, especially when local excitation is weak or candidate terms are correlated. To preserve a well-defined posterior solve, an adaptive information floor is used:
\begin{equation}
\rho_{\mathcal C,t}
=
\max
\left\{
0,
\epsilon_J
-
\lambda_{\min}
\left(
\boldsymbol J_{\mathcal C,t}^{0}
\right)
\right\},
\qquad
\epsilon_J>0,
\label{eq:method_information_floor}
\end{equation}
\begin{equation}
\boldsymbol J_{\mathcal C,t}
=
\boldsymbol J_{\mathcal C,t}^{0}
+
\rho_{\mathcal C,t}\boldsymbol I.
\label{eq:method_regularized_information}
\end{equation}
The candidate-subspace posterior is then
\begin{equation}
\boldsymbol\Xi_{\mathcal C,t}
=
\boldsymbol J_{\mathcal C,t}^{-1},
\qquad
\boldsymbol\mu_{\mathcal C,t}
=
\boldsymbol\Xi_{\mathcal C,t}
\boldsymbol h_{\mathcal C,t}^{0}.
\label{eq:method_candidate_posterior}
\end{equation}

The positive definiteness of the corrected information matrix follows directly from Eq.~\eqref{eq:method_information_floor}. If
\(\lambda_{\min}(\boldsymbol J_{\mathcal C,t}^{0})\ge\epsilon_J\), then \(\rho_{\mathcal C,t}=0\) and
\(\lambda_{\min}(\boldsymbol J_{\mathcal C,t})\ge\epsilon_J\). Otherwise,
\(\rho_{\mathcal C,t}=\epsilon_J-\lambda_{\min}(\boldsymbol J_{\mathcal C,t}^{0})\), and all eigenvalues of \(\boldsymbol J_{\mathcal C,t}^{0}\) are shifted by \(\rho_{\mathcal C,t}\). Hence
\begin{equation}
\lambda_{\min}(\boldsymbol J_{\mathcal C,t})
=
\epsilon_J,
\end{equation}
and therefore
\begin{equation}
\boldsymbol J_{\mathcal C,t}\succeq \epsilon_J\boldsymbol I,
\qquad
\|\boldsymbol J_{\mathcal C,t}^{-1}\|_2\le \epsilon_J^{-1}.
\label{eq:method_information_bound}
\end{equation}
This correction is not used to claim that each information increment is positive. It only ensures that the posterior solve remains well posed after adding new samples, discounting history, and removing expired samples.

In implementation, the Kronecker form is useful for deriving the posterior but does not need to be constructed explicitly. When the output residual covariance is diagonal or approximately diagonal after normalization,
\begin{equation}
\boldsymbol\Sigma_e
=
\operatorname{diag}
(\sigma_1^2,\ldots,\sigma_{n_y}^2),
\label{eq:method_diagonal_covariance}
\end{equation}
the candidate-subspace posterior mean can be obtained by solving the output channels separately. Let
\begin{equation}
\boldsymbol G_{\mathcal C,t}^{L}
=
\boldsymbol P_{\mathcal C_t}^{\top}
\boldsymbol G_t^{L}
\boldsymbol P_{\mathcal C_t},
\qquad
\boldsymbol Z_{\mathcal C,t}^{L}
=
\boldsymbol P_{\mathcal C_t}^{\top}
\boldsymbol Z_t^{L},
\end{equation}
and let \(\boldsymbol z_{\mathcal C,q,t}\) be the \(q\)-th column of
\(\boldsymbol Z_{\mathcal C,t}^{L}\). Define
\begin{equation}
\boldsymbol\Lambda_{\mathcal C,t}
=
\boldsymbol P_{\mathcal C_t}^{\top}
\boldsymbol\Lambda_t
\boldsymbol P_{\mathcal C_t}.
\end{equation}
Then the posterior mean for output channel \(q\) satisfies
\begin{equation}
\left(
\sigma_q^{-2}
\boldsymbol G_{\mathcal C,t}^{L}
+
\boldsymbol\Lambda_{\mathcal C,t}
\right)
\boldsymbol m_{\mathcal C,q,t}
=
\sigma_q^{-2}
\boldsymbol z_{\mathcal C,q,t},
\qquad
q=1,\ldots,n_y .
\label{eq:method_channelwise_solve}
\end{equation}
Equivalently,
\begin{equation}
\left(
\boldsymbol G_{\mathcal C,t}^{L}
+
\sigma_q^{2}
\boldsymbol\Lambda_{\mathcal C,t}
\right)
\boldsymbol m_{\mathcal C,q,t}
=
\boldsymbol z_{\mathcal C,q,t}.
\label{eq:method_channelwise_solve_equiv}
\end{equation}

This channel-wise form gives the same posterior mean as the diagonal-covariance Kronecker formulation. Its role is to avoid explicit construction and factorization of a \((|\mathcal C_t|n_y)\)-dimensional information matrix. In the present method, it should be understood as a practical implementation of the candidate-subspace posterior solve, rather than as a separate mechanism responsible for the identification performance.

\subsection{Posterior-guided shrinkage and support maintenance}
\label{subsec:posterior_guided_support}

The candidate-subspace posterior in Eq.~\eqref{eq:method_candidate_posterior} is a continuous posterior distribution. 
It gives the current coefficient mean and covariance, but it does not directly define the sparse model reported at time \(t\). 
This subsection converts the candidate posterior into three maintained objects: the shrinkage precision used in the next posterior update, the active support reported as the sparse model, and the candidate set used in the next candidate-subspace solve. 
The main idea is to use posterior evidence to refresh the horseshoe-induced shrinkage, then use a separate active-set projection to obtain an exactly sparse reported model.

For \(j\in\mathcal C_t\), let \(\boldsymbol E_{j,t}\) be the selection matrix that extracts from \(\boldsymbol\mu_{\mathcal C,t}\) the coefficient group associated with the \(j\)-th dictionary term. 
Define
\begin{equation}
\boldsymbol m_{j,t}
=
\boldsymbol E_{j,t}\boldsymbol\mu_{\mathcal C,t},
\qquad
\boldsymbol V_{j,t}
=
\boldsymbol E_{j,t}
\boldsymbol\Xi_{\mathcal C,t}
\boldsymbol E_{j,t}^{\top},
\label{eq:method_group_posterior_moments}
\end{equation}
where \(\boldsymbol m_{j,t}\in\mathbb R^{n_y}\) is the posterior mean of the coefficient group and \(\boldsymbol V_{j,t}\in\mathbb R^{n_y\times n_y}\) is its posterior covariance. 
The posterior second moment of this group is
\begin{equation}
q_{j,t}
=
\mathbb E
\left[
\|\boldsymbol b_{j,t}\|_2^2
\mid
\bar{\mathcal D}_t
\right]
=
\|\boldsymbol m_{j,t}\|_2^2
+
\operatorname{tr}
(\boldsymbol V_{j,t}).
\label{eq:method_group_second_moment}
\end{equation}
This quantity summarizes the posterior evidence carried by the \(j\)-th dictionary term. 
A small value of \(q_{j,t}\) indicates that the term has weak posterior support, while a persistent large value indicates that the term should be less strongly shrunk.

The shrinkage precision is refreshed from the same group-horseshoe prior introduced in Eq.~\eqref{eq:method_group_horseshoe}, rather than from an additional tuning rule. To make the refresh explicit, use the standard inverse-gamma augmentation of the half-Cauchy local and global scales. With the inverse-gamma density parameterized as \(x^{-a-1}\exp(-b/x)\), the conditional posterior of the local variance is
\begin{equation}
\lambda_{j,t}^{2}
\mid
\boldsymbol b_{j,t},\tau_t,\nu_{j,t}
\sim
\operatorname{IG}
\left(
\frac{n_y+1}{2},
\frac{1}{\nu_{j,t}}
+
\frac{\|\boldsymbol b_{j,t}\|_2^2}{2\tau_t^2}
\right).
\label{eq:method_local_scale_conditional}
\end{equation}
Since \(x\sim\operatorname{IG}(a,b)\) gives \(\mathbb E(x^{-1})=a/b\), the inverse local scale can be refreshed by replacing the unknown group energy \(\|\boldsymbol b_{j,t}\|_2^2\) with the posterior moment \(q_{j,t}\). Define
\begin{equation}
r_{j,t}
=
\lambda_{j,t}^{-2},
\qquad
g_t
=
\tau_t^{-2},
\qquad
\lambda^{\rm sh}_{j,t}
=
g_t r_{j,t}.
\label{eq:method_inverse_scale_def}
\end{equation}
Using the auxiliary inverse-scale constant \(\bar\nu_{j,t}^{-1}\) inherited from the previous local-scale update, the one-step empirical-Bayes local refresh is
\begin{equation}
\widetilde r_{j,t+1}
=
\frac{(n_y+1)/2}
{\bar\nu_{j,t}^{-1}
+
\frac{1}{2}g_t q_{j,t}},
\qquad
j\in\mathcal C_t .
\label{eq:method_local_scale_refresh}
\end{equation}
The same argument applied to the global scale gives
\begin{equation}
\widetilde g_{t+1}
=
\frac{(|\mathcal C_t|n_y+1)/2}
{\bar\zeta_t^{-1}
+
\frac{1}{2}
\sum_{j\in\mathcal C_t}
\widetilde r_{j,t+1}q_{j,t}},
\label{eq:method_global_scale_refresh}
\end{equation}
where \(\bar\nu_{j,t}^{-1}\) and \(\bar\zeta_t^{-1}\) denote the auxiliary inverse-scale constants carried over from the previous local and global scale updates, respectively; in implementation they are updated together with the refreshed scale estimates. 
Equations~\eqref{eq:method_local_scale_refresh} and \eqref{eq:method_global_scale_refresh} are therefore not additional tuning rules; they are posterior-mean refreshes of the inverse local and global scales under the conditional horseshoe hierarchy, with \(q_{j,t}\) used as a plug-in estimate of the current group coefficient energy.

The next shrinkage precision is obtained by clipping and damping:
\begin{equation}
\lambda^{\rm sh}_{j,t+1}
=
(1-\kappa_{\lambda})
\lambda^{\rm sh}_{j,t}
+
\kappa_{\lambda}
\operatorname{clip}
\left(
\widetilde g_{t+1}\widetilde r_{j,t+1},
\lambda_{\min}^{\rm sh},
\lambda_{\max}^{\rm sh}
\right),
\qquad
j\in\mathcal C_t .
\label{eq:method_shrinkage_refresh}
\end{equation}
For \(j\notin\mathcal C_t\), the previous shrinkage precision is retained. 
If such a term is reintroduced into \(\mathcal C_{t+1}\), the retained value is used as its initial precision in the next posterior solve. 
The clipping bounds prevent nearly unregularized directions and irreversible over-shrinkage, while the damping factor \(\kappa_{\lambda}\in(0,1]\) reduces abrupt support oscillations.

The continuous shrinkage update does not usually produce exact zero coefficients. 
The reported sparse support is therefore determined by a posterior relevance score:
\begin{equation}
s_{j,t}
=
\frac{
\|\boldsymbol m_{j,t}\|_2
}{
\sqrt{
\operatorname{tr}(\boldsymbol V_{j,t})+\varepsilon
}
},
\qquad
j\in\mathcal C_t,
\label{eq:method_posterior_relevance}
\end{equation}
where \(\varepsilon>0\) avoids division by zero. 
This score compares the posterior magnitude of a coefficient group with its posterior uncertainty. 
The preliminary active set is updated by the hysteresis rule
\begin{equation}
\mathcal S_t^{0}
=
\left\{
j\in\mathcal C_t:
s_{j,t}\ge \theta_{\rm on}
\quad
\text{or}
\quad
\left(
j\in\mathcal S_{t-1},
\ s_{j,t}\ge \theta_{\rm off}
\right)
\right\},
\qquad
0<\theta_{\rm off}<\theta_{\rm on}.
\label{eq:method_support_hysteresis}
\end{equation}
The higher threshold \(\theta_{\rm on}\) prevents noisy terms from entering the support, while the lower threshold \(\theta_{\rm off}\) avoids removing previously active terms too aggressively. 
Given the sparsity budget \(s_{\max}\), with \(s_{\max}\le C_{\max}\), the reported active support is
\begin{equation}
\mathcal S_t
=
\begin{cases}
\mathcal S_t^{0},
&
|\mathcal S_t^{0}|\le s_{\max},
\\[1mm]
\operatorname{Top}_{s_{\max}}
(\mathcal S_t^{0};s_{j,t}),
&
|\mathcal S_t^{0}|>s_{\max}.
\end{cases}
\label{eq:method_support_budget}
\end{equation}

The sparse coefficient matrix reported at time \(t\) is obtained by projecting the posterior mean onto \(\mathcal S_t\):
\begin{equation}
\widehat{\boldsymbol b}_{j,t}^{\top}
=
\begin{cases}
\boldsymbol m_{j,t}^{\top}, & j\in\mathcal S_t,\\
\boldsymbol 0^{\top}, & j\notin\mathcal S_t.
\end{cases}
\label{eq:method_sparse_projection}
\end{equation}
The unprojected posterior mean is still retained for uncertainty evaluation and for candidate-set refresh, whereas \(\widehat{\boldsymbol B}_t\) is the compact model used for support reporting and sparsity evaluation.

Finally, residual information is used to allow inactive terms to re-enter when the operating condition changes. 
Let \(\mathcal W_t\) be a short recent window and define
\begin{equation}
\boldsymbol R_{\mathcal W_t}
=
\bar{\boldsymbol Y}_{\mathcal W_t}
-
\boldsymbol\Psi_{\mathcal W_t}
\boldsymbol B_t^{\rm mean},
\label{eq:method_short_residual}
\end{equation}
where \(\boldsymbol B_t^{\rm mean}\in\mathbb R^{p\times n_y}\) denotes the unprojected posterior mean embedded in the full dictionary, with zero rows outside \(\mathcal C_t\). 
For inactive terms, define the residual relevance score
\begin{equation}
\eta_{j,t}
=
\frac{
\left\|
\boldsymbol\psi_{j,\mathcal W_t}^{\top}
\boldsymbol R_{\mathcal W_t}
\right\|_2
}{
\|\boldsymbol\psi_{j,\mathcal W_t}\|_2+\varepsilon
},
\qquad
j\notin\mathcal S_t,
\label{eq:method_residual_relevance}
\end{equation}
where \(\boldsymbol\psi_{j,\mathcal W_t}\) is the column of \(\boldsymbol\Psi_{\mathcal W_t}\) corresponding to the \(j\)-th dictionary term. 
The next candidate set is then constructed as
\begin{equation}
\mathcal C_{t+1}
=
\mathcal S_t
\cup
\operatorname{Top}_{C_{\max}-|\mathcal S_t|}
\left(
\{j\notin\mathcal S_t\};\eta_{j,t}
\right).
\label{eq:method_candidate_refresh}
\end{equation}
Thus, \(\mathcal S_t\) defines the sparse model reported at the current time, while \(\mathcal C_{t+1}\) defines the computational subspace for the next posterior solve.
\subsection{Error relation under coefficient drift}
\label{subsec:bounded_tracking}

The recursive posterior update tracks a coefficient vector that may change slowly with the operating condition. This subsection gives a bounded-error relation for the candidate-subspace posterior mean and clarifies how the estimation error is affected by shrinkage, residual information, coefficient drift, and posterior conditioning.

Let \(\boldsymbol\beta_{\mathcal C,t}^{\star}\) denote the ideal coefficient vector restricted to the current candidate subspace. Define the likelihood information matrix on \(\mathcal C_t\) as
\begin{equation}
\boldsymbol H_{\mathcal C,t}^{L}
=
\boldsymbol T_{\mathcal C_t}^{\top}
\left(
\boldsymbol\Sigma_e^{-1}
\otimes
\boldsymbol G_t^{L}
\right)
\boldsymbol T_{\mathcal C_t}.
\label{eq:tracking_likelihood_information}
\end{equation}
The corresponding information vector can be decomposed as
\begin{equation}
\boldsymbol h_{\mathcal C,t}^{0}
=
\boldsymbol H_{\mathcal C,t}^{L}
\boldsymbol\beta_{\mathcal C,t}^{\star}
+
\boldsymbol r_{\mathcal C,t}^{L},
\label{eq:tracking_effective_information}
\end{equation}
where \(\boldsymbol r_{\mathcal C,t}^{L}\) collects the residual information caused by measurement noise, unmodeled dictionary contributions, and the mismatch between older window samples and the current coefficient vector.

Suppose that the coefficient drift satisfies
\begin{equation}
\boldsymbol B_t^{\star}
=
\boldsymbol B_{t-1}^{\star}
+
\boldsymbol\Omega_t,
\qquad
\|\boldsymbol\Omega_t\|_F\le \delta .
\label{eq:tracking_drift}
\end{equation}
An \(h\)-step-old sample may then carry a coefficient mismatch of order \(h\delta\). The accumulated effect over the sliding window is represented by
\begin{equation}
\Gamma_L(\xi)
=
\left(
\sum_{h=0}^{L-1}
\xi^{h}h^2
\right)^{1/2},
\label{eq:tracking_gamma}
\end{equation}
which increases with the effective window length and decreases when older samples are more strongly discounted.

Assume that there exist constants \(\bar B>0\), \(\bar E>0\), \(\bar\phi_\Sigma>0\), \(\bar\psi_d>0\), \(\bar\rho>0\), \(\lambda_{\max}^{\rm sh}>0\), and \(\epsilon_J>0\) such that
\begin{equation}
\|\boldsymbol\beta_{\mathcal C,t}^{\star}\|_2\le \bar B,
\qquad
\lambda_{\min}(\boldsymbol J_{\mathcal C,t})\ge \epsilon_J,
\qquad
0\le \rho_{\mathcal C,t}\le \bar\rho,
\label{eq:tracking_basic_bounds}
\end{equation}
\begin{equation}
\|\boldsymbol\Lambda_{\mathcal C,t}\|_2
\le
\lambda_{\max}^{\rm sh},
\qquad
\|\boldsymbol r_{\mathcal C,t}^{L}\|_2
\le
\bar\phi_\Sigma
\left(
\bar E
+
\bar\psi_d
\Gamma_L(\xi)\delta
\right),
\label{eq:tracking_residual_bound}
\end{equation}
where \(\bar E\) represents the nominal residual level, and \(\bar\psi_d\) absorbs the influence of bounded dictionary functions and covariance weighting. Under these conditions, the candidate-subspace posterior mean satisfies
\begin{equation}
\left\|
\boldsymbol\mu_{\mathcal C,t}
-
\boldsymbol\beta_{\mathcal C,t}^{\star}
\right\|_2
\le
\frac{
(\lambda_{\max}^{\rm sh}+\bar\rho)\bar B
+
\bar\phi_\Sigma
\left(
\bar E
+
\bar\psi_d\Gamma_L(\xi)\delta
\right)
}{
\epsilon_J
}.
\label{eq:tracking_bound_revised}
\end{equation}

To prove Eq.~\eqref{eq:tracking_bound_revised}, write the regularized posterior information matrix as
\begin{equation}
\boldsymbol J_{\mathcal C,t}
=
\boldsymbol H_{\mathcal C,t}^{L}
+
\boldsymbol\Lambda_{\mathcal C,t}
+
\rho_{\mathcal C,t}\boldsymbol I.
\label{eq:tracking_J_decomposition}
\end{equation}
Using Eq.~\eqref{eq:tracking_effective_information}, the posterior mean is
\begin{align}
\boldsymbol\mu_{\mathcal C,t}
&=
\boldsymbol J_{\mathcal C,t}^{-1}
\boldsymbol h_{\mathcal C,t}^{0} \nonumber\\
&=
\boldsymbol J_{\mathcal C,t}^{-1}
\left(
\boldsymbol H_{\mathcal C,t}^{L}
\boldsymbol\beta_{\mathcal C,t}^{\star}
+
\boldsymbol r_{\mathcal C,t}^{L}
\right).
\end{align}
Since
\[
\boldsymbol H_{\mathcal C,t}^{L}
=
\boldsymbol J_{\mathcal C,t}
-
\boldsymbol\Lambda_{\mathcal C,t}
-
\rho_{\mathcal C,t}\boldsymbol I,
\]
we obtain
\begin{equation}
\boldsymbol\mu_{\mathcal C,t}
-
\boldsymbol\beta_{\mathcal C,t}^{\star}
=
-
\boldsymbol J_{\mathcal C,t}^{-1}
\left(
\boldsymbol\Lambda_{\mathcal C,t}
+
\rho_{\mathcal C,t}\boldsymbol I
\right)
\boldsymbol\beta_{\mathcal C,t}^{\star}
+
\boldsymbol J_{\mathcal C,t}^{-1}
\boldsymbol r_{\mathcal C,t}^{L}.
\end{equation}
Taking norms gives
\begin{align}
\left\|
\boldsymbol\mu_{\mathcal C,t}
-
\boldsymbol\beta_{\mathcal C,t}^{\star}
\right\|_2
&\le
\|\boldsymbol J_{\mathcal C,t}^{-1}\|_2
\left[
\left\|
\boldsymbol\Lambda_{\mathcal C,t}
+
\rho_{\mathcal C,t}\boldsymbol I
\right\|_2
\|\boldsymbol\beta_{\mathcal C,t}^{\star}\|_2
+
\|\boldsymbol r_{\mathcal C,t}^{L}\|_2
\right] \nonumber\\
&\le
\frac{
(\lambda_{\max}^{\rm sh}+\bar\rho)\bar B
+
\bar\phi_\Sigma
\left(
\bar E
+
\bar\psi_d\Gamma_L(\xi)\delta
\right)
}{
\epsilon_J
}.
\end{align}
This proves Eq.~\eqref{eq:tracking_bound_revised}.

The complete online procedure is summarized in Algorithm~\ref{alg:online_brsl}. The main recursive state consists of the sliding-window statistics, the shrinkage precision, the active support, and the candidate set.

\begin{algorithm}[H]
\caption{Online BRSL with sliding-window likelihood-ratio information recursion and support maintenance}
\label{alg:online_brsl}
\begin{algorithmic}[1]
\REQUIRE Normalized dictionary \(\boldsymbol\psi(\cdot)\), window length \(L\), forgetting factor \(\xi\), maximum candidate size \(C_{\max}\), sparsity budget \(s_{\max}\), thresholds \(\theta_{\rm on}\), \(\theta_{\rm off}\), damping factor \(\kappa_\lambda\), precision bounds \(\lambda_{\min}^{\rm sh},\lambda_{\max}^{\rm sh}\), residual window \(\mathcal W_t\), and information floor \(\epsilon_J\), with \(s_{\max}\le C_{\max}\).
\STATE Initialize \(\boldsymbol G_0^{L}=\boldsymbol 0\), \(\boldsymbol Z_0^{L}=\boldsymbol 0\), shrinkage precision \(\boldsymbol\Lambda_1\), active support \(\mathcal S_0\), and candidate set \(\mathcal C_1\).
\FOR{\(t=1,2,\ldots\)}
    \STATE Receive new data and determine the added set \(\mathcal A_t\) and expired set \(\mathcal R_t\).
    \STATE Update \(\boldsymbol G_t^{L}\) and \(\boldsymbol Z_t^{L}\) using Eqs.~\eqref{eq:method_G_recursion} and \eqref{eq:method_Z_recursion}.
    \STATE Form \(\boldsymbol J_t^{L,0}\) and \(\boldsymbol h_t^{L,0}\) using Eqs.~\eqref{eq:method_raw_full_J} and \eqref{eq:method_raw_full_h}.
    \STATE Restrict the posterior information to \(\mathcal C_t\) using Eq.~\eqref{eq:method_candidate_information}.
    \STATE Apply the adaptive information floor using Eqs.~\eqref{eq:method_information_floor} and \eqref{eq:method_regularized_information}.
    \STATE Solve the candidate-subspace posterior using Eq.~\eqref{eq:method_candidate_posterior}; when \(\boldsymbol\Sigma_e\) is diagonal, use the equivalent channel-wise solve in Eq.~\eqref{eq:method_channelwise_solve} to avoid explicit Kronecker construction.
    \STATE Compute group posterior moments \(q_{j,t}\) using Eq.~\eqref{eq:method_group_second_moment}.
    \STATE Refresh the shrinkage precision \(\boldsymbol\Lambda_{t+1}\) using Eqs.~\eqref{eq:method_local_scale_refresh}--\eqref{eq:method_shrinkage_refresh}.
    \STATE Compute posterior relevance scores \(s_{j,t}\) and update the active support \(\mathcal S_t\) using Eqs.~\eqref{eq:method_posterior_relevance}--\eqref{eq:method_support_budget}.
    \STATE Obtain the sparse reported coefficients by the projection in Eq.~\eqref{eq:method_sparse_projection}.
    \STATE Compute residual relevance scores \(\eta_{j,t}\) and construct \(\mathcal C_{t+1}\) using Eqs.~\eqref{eq:method_residual_relevance} and \eqref{eq:method_candidate_refresh}.
\ENDFOR
\end{algorithmic}
\end{algorithm}

\section{Numerical simulations}\label{sec4}

\subsection{Experiment 1: Sparse coefficient tracking under known support}
\label{subsec:exp1}

The first experiment uses a controlled nonlinear regression problem with known time-varying sparse coefficients. The purpose is to evaluate whether the proposed method can recursively track drifting coefficients, recover the active dictionary support, and suppress inactive terms under an over-complete dictionary. Since the true coefficient matrix is known, this experiment allows direct evaluation of prediction accuracy, coefficient tracking accuracy, and support recovery.

The simulated system has three regressors, denoted by \(x_1\), \(x_2\), and \(x_3\), and two normalized outputs, denoted by \(y_1\) and \(y_2\). A candidate dictionary with \(p=80\) terms is used. The true group support contains six active dictionary terms:
\[
\mathcal S^\star
=
\{4,5,8,13,19,24\},
\]
corresponding to
\[
x_1^2,\quad
x_2^2,\quad
x_1x_3,\quad
x_1x_2x_3,\quad
x_1^2x_2,\quad
x_1^2x_2x_3 .
\]
The same active dictionary terms are used for both outputs, but the coefficient trajectories are output-dependent. Therefore, the coefficient tracking results are reported separately for \(y_1\) and \(y_2\). The true coefficients vary slowly over time, with a drift amplitude set to \(15\%\) of the corresponding nominal coefficient magnitude. Additive Gaussian noise with standard deviation \(0.06\) is added to the normalized outputs.

The total data length is \(T=2200\). The first \(N_0=350\) samples are used for initial calibration, and the remaining samples are used for online recursive evaluation. The proposed method uses a sliding window length \(L=150\), forgetting factor \(\xi=0.992\), maximum candidate-set size \(C_{\max}=36\), maximum active support size \(s_{\max}=12\), and support threshold \(\varepsilon_s=0.06\). The compared methods include a parameter-tracking Kalman filter (Param-KF), sliding-window sparse Bayesian learning (SW-SBL), and sliding-window least absolute shrinkage and selection operator (LASSO; SW-LASSO). All methods use the same normalized dictionary and the same online data stream.

\begin{center}
\safeincludegraphics[width=0.90\linewidth,height=0.45\textheight,keepaspectratio]{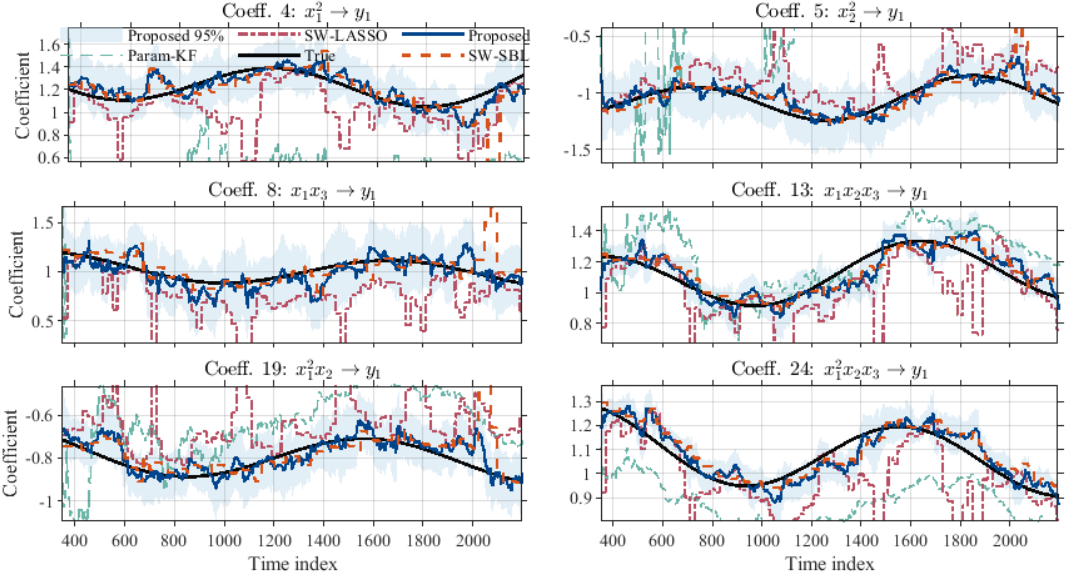}
\captionof{figure}{Coefficient tracking results for \(y_1\) in Experiment~1. The six panels correspond to the six true active dictionary terms. The shaded region denotes the proposed method's coefficient-wise \(95\%\) posterior interval.}
\label{fig:exp1_coeff_y1}
\end{center}

Fig.~\ref{fig:exp1_coeff_y1} shows the coefficient tracking results for \(y_1\). The proposed method follows the main drift trend of all six true active coefficients and remains close to the ground truth over most of the online period. The posterior interval also reflects local uncertainty variation during recursive updating. The comparison methods are shown using light dashed curves. Param-KF and SW-SBL can sometimes follow the slow trend, but their estimates show more visible deviations or spurious fluctuations. SW-LASSO is less stable, especially when correlated inactive terms compete with the true active terms.

\begin{center}
\safeincludegraphics[width=0.90\linewidth,height=0.45\textheight,keepaspectratio]{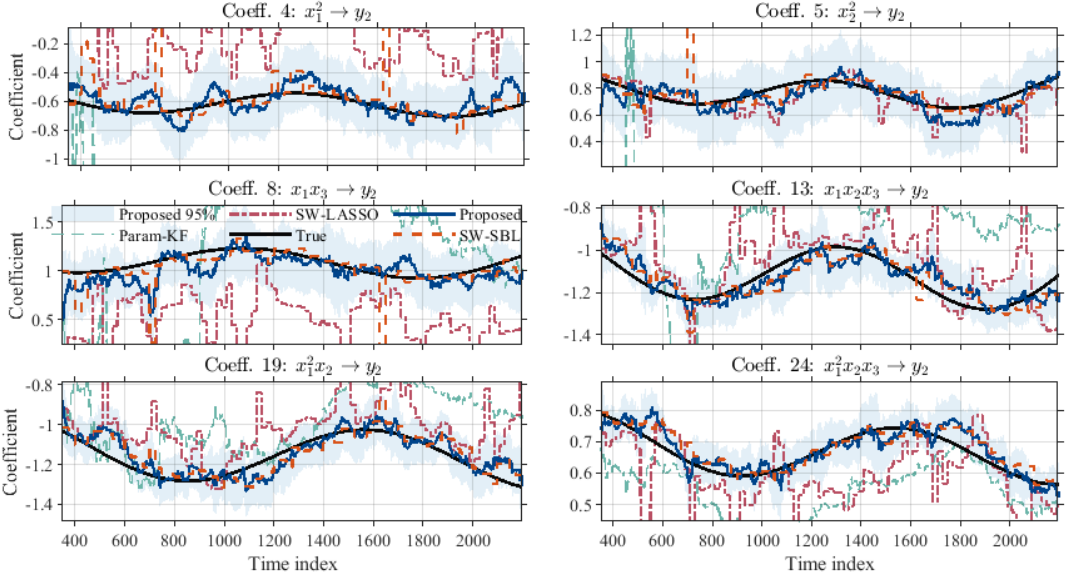}
\captionof{figure}{Coefficient tracking results for \(y_2\) in Experiment~1. The same six active dictionary terms are used as in Fig.~\ref{fig:exp1_coeff_y1}, but the coefficient trajectories are output-dependent.}
\label{fig:exp1_coeff_y2}
\end{center}

Fig.~\ref{fig:exp1_coeff_y2} gives the corresponding coefficient tracking results for \(y_2\). The same six dictionary terms are active, but their coefficient values and drift trajectories differ from those in \(y_1\). The proposed method again provides the most stable tracking behavior among the compared sparse methods. These two figures show that the proposed method does not merely identify a common support; it also tracks output-dependent coefficient trajectories within the same group-sparse dictionary structure.

\paragraph{Evaluation metrics.}
Let \(\mathcal T_{\mathrm{on}}\) be the online evaluation period, with \(N_{\mathrm{on}}=|\mathcal T_{\mathrm{on}}|\). For method \(m\), the one-step prediction, coefficient estimate, and true coefficient matrix are denoted by \(\widehat{\boldsymbol y}_{t}^{(m)}\), \(\widehat{\boldsymbol B}_{t}^{(m)}\), and \(\boldsymbol B_t^\star\), respectively. The active set \(\widehat{\mathcal S}_{t}^{(m)}\) is obtained by applying the same row-wise group-norm threshold \(\varepsilon_s\) to all methods; for the proposed method this coincides with the posterior-projected active set.

Prediction accuracy is measured by the root-mean-square error (RMSE), and coefficient-tracking accuracy is measured by
\begin{equation}
\label{eq:exp1_rmse_metric}
\begin{aligned}
\mathrm{RMSE}^{(m)}
&=\left[\frac{1}{N_{\mathrm{on}}}\sum_{t\in\mathcal T_{\mathrm{on}}}
\|\boldsymbol y_t-\widehat{\boldsymbol y}_{t}^{(m)}\|_2^2\right]^{1/2},\\
\mathrm{CoefErr}^{(m)}
&=\frac{1}{N_{\mathrm{on}}}\sum_{t\in\mathcal T_{\mathrm{on}}}
\frac{\|\widehat{\boldsymbol B}_{t}^{(m)}-\boldsymbol B_t^\star\|_F}{\|\boldsymbol B_t^\star\|_F},\\
\mathrm{OnErr}^{(m)}
&=\frac{1}{N_{\mathrm{on}}}\sum_{t\in\mathcal T_{\mathrm{on}}}
\frac{\|\widehat{\boldsymbol B}_{t,\mathcal S^\star}^{(m)}-\boldsymbol B_{t,\mathcal S^\star}^{\star}\|_F}{\|\boldsymbol B_{t,\mathcal S^\star}^{\star}\|_F},
\end{aligned}
\end{equation}
where \(\mathcal S^\star\) is the true support. Model leakage and support compactness are evaluated by
\begin{equation}
\label{eq:exp1_support_metrics}
\begin{aligned}
\mathrm{OffE}^{(m)}
&=\frac{1}{N_{\mathrm{on}}}\sum_{t\in\mathcal T_{\mathrm{on}}}
\frac{\|\widehat{\boldsymbol B}_{t,\bar{\mathcal S}^{\star}}^{(m)}\|_F}{\|\boldsymbol B_{t,\mathcal S^\star}^{\star}\|_F},\\
\mathrm{F1}^{(m)}
&=\frac{1}{N_{\mathrm{on}}}\sum_{t\in\mathcal T_{\mathrm{on}}}
\frac{2\mathrm{Prec}_t^{(m)}\mathrm{Rec}_t^{(m)}}{\mathrm{Prec}_t^{(m)}+\mathrm{Rec}_t^{(m)}+\epsilon},\\
\mathrm{ActSize}^{(m)}
&=\frac{1}{N_{\mathrm{on}}}\sum_{t\in\mathcal T_{\mathrm{on}}}|\widehat{\mathcal S}_{t}^{(m)}|,
\end{aligned}
\end{equation}
where \(\bar{\mathcal S}^{\star}\) is the complement of the true support, and \(\epsilon\) avoids division by zero. Precision and recall are computed from \(\widehat{\mathcal S}_{t}^{(m)}\) and \(\mathcal S^\star\) in the standard way. The rolling RMSE shown below uses a 60-sample window.

\begin{center}
\safeincludegraphics[width=0.90\linewidth,height=0.45\textheight,keepaspectratio]{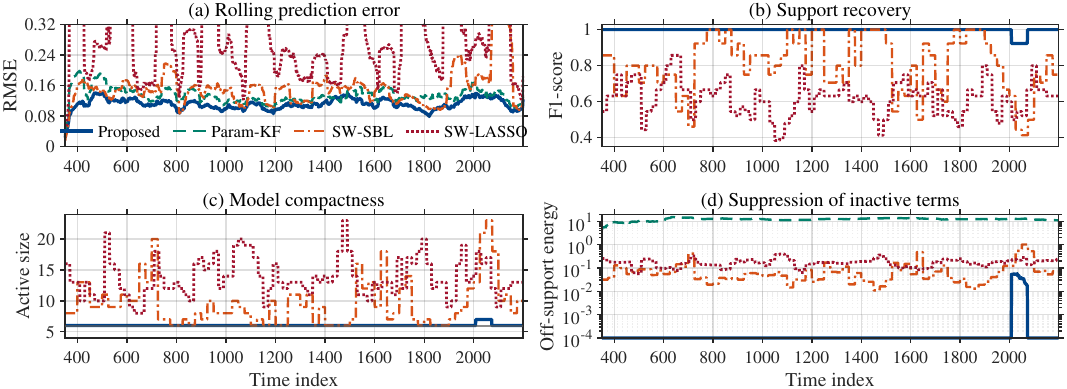}
\captionof{figure}{Online performance diagnostics in Experiment~1: (a) rolling prediction RMSE; (b) support-recovery F1-score; (c) number of active dictionary terms; and (d) off-support coefficient energy.}
\label{fig:exp1_diagnostics}
\end{center}

Fig.~\ref{fig:exp1_diagnostics} compares the online prediction and sparsity diagnostics. In Fig.~\ref{fig:exp1_diagnostics}(a), the proposed method maintains a low and stable rolling RMSE over the online period. Param-KF also gives competitive prediction in some intervals, but this is achieved by updating a dense coefficient vector. The sparse baselines show larger fluctuations, especially SW-LASSO. Fig.~\ref{fig:exp1_diagnostics}(b) shows that the proposed method maintains an F1-score close to one for most of the online period, indicating stable recovery of the true group support. In Fig.~\ref{fig:exp1_diagnostics}(c), the proposed method keeps the active size close to the true support size of six, while SW-SBL and SW-LASSO select larger and more fluctuating supports. Fig.~\ref{fig:exp1_diagnostics}(d) further shows that the proposed method suppresses inactive dictionary terms effectively, whereas Param-KF keeps a much larger off-support coefficient energy.

\begin{center}
\captionof{table}{Average online performance in Experiment~1.}
\label{tab:exp1_metrics}
\small
\setlength{\tabcolsep}{3.5pt}
\renewcommand{\arraystretch}{0.95}
\begin{tabular}{lcccccc}
\hline
Method
& RMSE
& CoefErr
& OnErr
& OffE
& F1
& ActSize \\
\hline
Proposed
& 0.1138
& 0.0730
& 0.0727
& 0.0014
& 0.9973
& 6.04 \\
Param-KF
& 0.1368
& 12.1664
& 1.4426
& 12.0754
& 0.1421
& 78.47 \\
SW-SBL
& 0.1690
& 0.1279
& 0.0798
& 0.0952
& 0.8062
& 9.59 \\
SW-LASSO
& 0.4254
& 0.3310
& 0.2810
& 0.1746
& 0.6196
& 13.78 \\
\hline
\end{tabular}
\end{center}

Table~\ref{tab:exp1_metrics} summarizes the average online metrics. The proposed method achieves the lowest prediction RMSE, the smallest coefficient tracking error, and the lowest off-support energy. Its average active size is \(6.04\), which is very close to the true support size of six. This shows that the proposed method does not obtain good prediction by using a dense model; instead, it preserves a compact and accurate support while tracking drifting coefficients.

Param-KF has a moderate RMSE but a very large coefficient error and off-support energy, which indicates that dense adaptive estimation can fit the output without recovering the underlying sparse structure. SW-SBL improves sparsity compared with Param-KF, but it still selects extra inactive terms and gives a lower F1-score than the proposed method. SW-LASSO has the largest RMSE and a fluctuating active support, showing that deterministic sliding-window shrinkage is less stable in this drifting and correlated-dictionary setting.

Overall, Experiment~1 verifies the intended role of the proposed method in a controlled setting. With an over-complete dictionary and known drifting sparse coefficients, the proposed method provides accurate online prediction, stable coefficient tracking, reliable support recovery, and effective suppression of inactive dictionary terms.

\FloatBarrier

\subsection{Experiment 2: Power-plant-oriented nonlinear time-varying MIMO identification}
\label{sec:exp2}

The second experiment uses a power-plant-oriented nonlinear benchmark to examine whether the proposed recursive sparse Bayesian scheme can maintain a compact time-varying model under operating-condition changes. Different from Experiment~1, the output is a six-channel derivative vector generated by a recursive dynamic system. The benchmark is constructed around a 600-MW subcritical boiler--turbine--pulverizing process, so that the variables and input ranges have engineering meanings, while the true sparse coefficients remain available for quantitative support evaluation.

The physical variables are first converted into normalized deviation variables. The six states are drum pressure, electric power, drum-level deviation, raw-coal inventory, pulverized-coal inventory, and mill outlet temperature. They are denoted by
\begin{equation}
    \boldsymbol{x}_k
    =
    [p_{d,k},\; P_{e,k},\; z_{d,k},\; M_{c,k},\; M_{pf,k},\; T_{m,k}]^{\top} .
    \label{eq:exp2_powerplant_state}
\end{equation}
The manipulated and disturbance-related inputs are coal-feed flow, turbine-valve opening, feedwater flow, primary-air flow, and primary-air temperature,
\begin{equation}
    \boldsymbol{u}_k
    =
    [F_{c,k},\; \alpha_{v,k},\; F_{w,k},\; F_{pa,k},\; T_{pa,k}]^{\top} .
    \label{eq:exp2_powerplant_input}
\end{equation}
All variables used for identification are normalized by nominal values and scales, while the generated trajectories are also stored in physical units for plotting. The load command is restricted to the 240--600 MW range, and the tested trajectory corresponds to a bounded load-following segment with small distributed-control-system-like (DCS-like) perturbations in the input channels.

In the normalized coordinate, the recursive data generator is written as
\begin{equation}
    \boldsymbol{x}_{k+1}
    =
    \boldsymbol{x}_{k}
    +
    \Delta t\,\boldsymbol{f}_{k}(\boldsymbol{x}_{k},\boldsymbol{u}_{k})
    +
    \boldsymbol{w}_{k},
    \label{eq:exp2_powerplant_recursion}
\end{equation}
where
\begin{equation}
    \boldsymbol{f}_{k}(\boldsymbol{x}_{k},\boldsymbol{u}_{k})
    =
    \boldsymbol{\Theta}^{\star}_{k}\boldsymbol{\phi}(\boldsymbol{x}_{k},\boldsymbol{u}_{k}) .
    \label{eq:exp2_powerplant_sparse_dynamics}
\end{equation}
The measured regression target is the noisy derivative vector
\begin{equation}
    \boldsymbol{y}_{k}
    =
    \boldsymbol{\Theta}^{\star}_{k}\boldsymbol{\phi}(\boldsymbol{x}_{k},\boldsymbol{u}_{k})
    +
    \boldsymbol{e}_{k} ,
    \label{eq:exp2_powerplant_regression}
\end{equation}
where \(\boldsymbol{w}_{k}\) is a small process disturbance and \(\boldsymbol{e}_{k}\) denotes derivative measurement noise. Disturbance bursts are also added to a small fraction of the samples to emulate abnormal measurement fluctuations.

The candidate dictionary contains 26 terms and is written directly with the normalized physical variables,
\begin{align}
\boldsymbol{\phi}(\boldsymbol{x},\boldsymbol{u})=
[&1,
 p_d, P_e, z_d, M_c, M_{pf}, T_m,
 F_c, \alpha_v, F_w, F_{pa}, T_{pa},
\notag\\
&p_dP_e, p_dM_{pf}, P_ez_d, M_cM_{pf}, M_{pf}T_m,
\notag\\
&p_d\alpha_v, p_dF_w, P_e\alpha_v, z_dF_w, M_cF_c,
 M_cF_{pa}, M_{pf}F_{pa}, T_mF_{pa}, T_mT_{pa}]^{\top} .
\label{eq:exp2_powerplant_dictionary}
\end{align}
The dictionary is not a full polynomial expansion. It keeps the main self-dynamic terms, input channels, inventory-related interactions, and state--input products that are plausible for a boiler--turbine--pulverizing process. This design makes the dictionary correlated enough to be challenging, while each selected term still has a clear physical interpretation.

The true coefficient matrix 
\(\boldsymbol{\Theta}^{\star}_{k}\in\mathbb{R}^{6\times 26}\) is sparse and channel dependent, where the rows correspond to the six output channels and the columns correspond to the dictionary terms. 
Let \(\mathcal J_q^{\star}\) denote the true active dictionary-index set of the \(q\)-th output channel. 
Only the coefficients associated with \(\mathcal J_q^{\star}\) are allowed to vary with time, while the remaining coefficients are set to zero. 
For \(j\in\mathcal J_q^{\star}\), the coefficient trajectory is generated as
\begin{equation}
    \theta^{\star}_{qj,k}
    =
    \bar{\theta}_{qj}
    \left[
    1+a_{qj}\sin(\omega_{qj}t_k+\varphi_{qj})
    \right]
    +
    \Delta\theta_{qj}\chi(t_k),
    \label{eq:exp2_powerplant_drift}
\end{equation}
where the sinusoidal term represents slow coefficient drift, and the second term introduces a smooth offset caused by the operating-condition change. 
The transition profile is defined as
\begin{equation}
    \chi(t_k)
    =
    \frac{1}{2}
    \left[
    1+
    \tanh
    \left(
    \frac{t_k-t_c}{\tau_c}
    \right)
    \right],
    \label{eq:exp2_transition_profile}
\end{equation}
where \(t_c\) denotes the operating-change time and \(\tau_c\) controls the transition width. 
For \(j\notin\mathcal J_q^{\star}\), \(\theta^{\star}_{qj,k}=0\). 
Thus, the benchmark contains both slow coefficient drift and a localized operating-regime transition, while retaining a fixed sparse support structure for evaluation.

The data length of the main test is \(T=900\), with sampling interval \(\Delta t=0.03\). 
The first \(N_0=110\) samples are used for normalization and initialization, and the remaining samples are processed sequentially. 
The proposed method uses a window length \(L=110\) and a forgetting factor \(\xi=0.992\). 
The rolling error is computed with \(M=36\) samples. 
Four baselines are considered: forgetting-factor recursive least squares (FF-RLS), parameter Kalman filtering (Param-KF), sliding-window LASSO (SW-LASSO), and sliding-window sparse Bayesian learning (SW-SBL). 
All methods use the same normalized dictionary, the same initial segment, and the same online data stream.

The rolling one-step prediction error is computed as
\begin{equation}
    \mathrm{RMSE}_{m,k}
    =
    \left[
    \frac{1}{Mn_y}
    \sum_{i=k-M+1}^{k}
    \|\hat{\boldsymbol{y}}_{m,i}-\boldsymbol{y}_{i}\|_2^2
    \right]^{1/2},
    \label{eq:exp2_powerplant_rmse}
\end{equation}
where \(m\) denotes the method. The online compactness is measured by the mean number of active dictionary terms per output channel. To describe coefficient leakage outside the true support, the relative off-support coefficient energy is defined as
\begin{equation}
    E^{\mathrm{off}}_{m,k}
    =
    \frac{
    \sum_{q=1}^{n_y}
    \sum_{j\notin S^{\star}_{q,k}}
    \hat\theta_{m,qj,k}^{2}
    }{
    \sum_{q=1}^{n_y}
    \sum_{j=1}^{p}
    \hat\theta_{m,qj,k}^{2}
    +\epsilon
    },
    \label{eq:exp2_powerplant_off_energy}
\end{equation}
where \(S^{\star}_{q,k}\) is the true active support of channel \(q\). For the proposed method, the one-step predictive variance is obtained from the posterior covariance,
\begin{equation}
    \hat s_{q,k}^{2}
    =
    \boldsymbol{\phi}_{k}^{\top}
    \boldsymbol{\Xi}_{q,k}
    \boldsymbol{\phi}_{k}
    +
    \hat\sigma_{q,k}^{2},
    \label{eq:exp2_powerplant_predvar}
\end{equation}
and the nominal \(95\%\) predictive interval is \(\hat y_{q,k}\pm1.96\hat s_{q,k}\).

\begin{center}
\safeincludegraphics[width=0.90\linewidth,height=0.42\textheight,keepaspectratio]{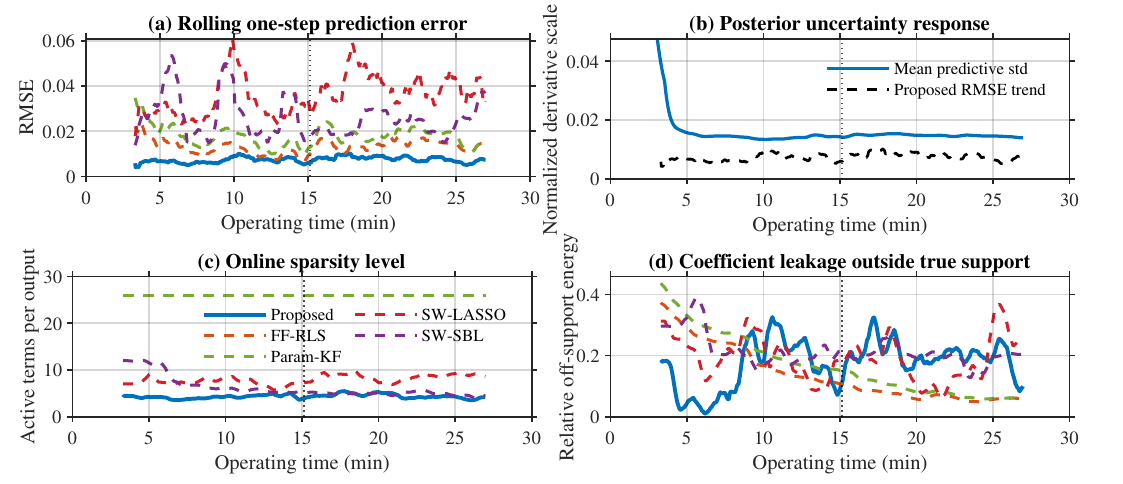}
\captionof{figure}{Online prediction, sparsity, and uncertainty diagnostics in the power-plant-oriented MIMO nonlinear time-varying identification test.
(a) Rolling one-step RMSE.
(b) Mean posterior predictive scale of the proposed method, with the rolling RMSE shown as a reference.
(c) Online sparsity level.
(d) Relative off-support coefficient energy.}
\label{fig:exp2_dynamic_accuracy}
\end{center}

Fig.~\ref{fig:exp2_dynamic_accuracy} summarizes the online behavior of different methods over the validation interval. 
As shown in Fig.~\ref{fig:exp2_dynamic_accuracy}(a), the proposed method maintains a low rolling one-step RMSE under changing operating conditions. 
The dense recursive baselines can adapt to the measurements, but their active dictionary sizes remain large, as shown in Fig.~\ref{fig:exp2_dynamic_accuracy}(c). 
In contrast, the proposed method keeps a compact active set while preserving prediction accuracy. 
Fig.~\ref{fig:exp2_dynamic_accuracy}(d) further shows that its coefficient energy is mainly concentrated on the true support, indicating that the online update does not rely on widespread off-support coefficients.

Fig.~\ref{fig:exp2_dynamic_accuracy}(b) reports the mean posterior predictive scale of the proposed method. 
The curve is relatively large during the early online adaptation stage and then settles to a lower level as local information accumulates. 
Its magnitude remains close to the rolling RMSE trend, showing that the posterior covariance provides a consistent global uncertainty scale for the online predictor. 
This result complements the error and sparsity curves by showing that the proposed method outputs both point predictions and posterior uncertainty information during recursive identification.

\begin{center}
\safeincludegraphics[width=0.90\linewidth,height=0.42\textheight,keepaspectratio]{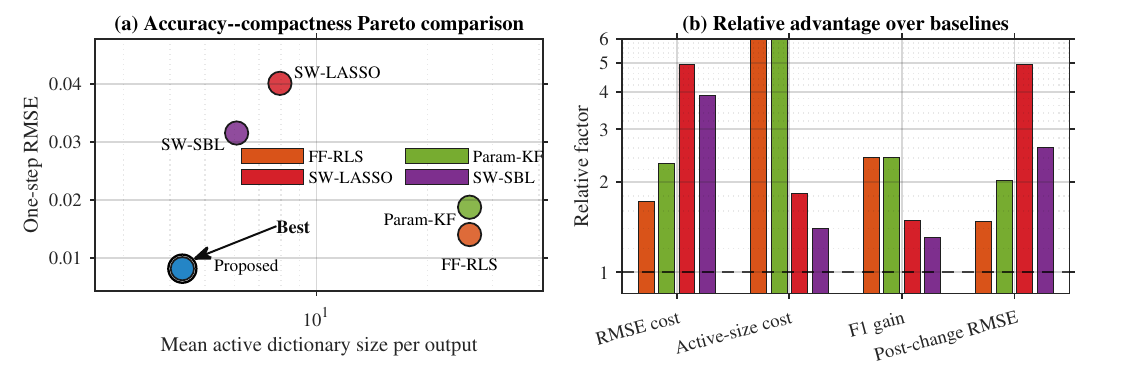}
\captionof{figure}{Summary comparison of prediction accuracy, model compactness, and relative metric ratios. 
(a) Accuracy--compactness Pareto comparison. The lower-left region is preferred. 
(b) Ratios of baseline metrics with respect to the proposed method.}
\label{fig:exp2_pareto}
\end{center}

Fig.~\ref{fig:exp2_pareto} summarizes the trade-off between prediction accuracy and compactness. In the Pareto map, the proposed method is located closest to the lower-left region. The average one-step RMSE of the proposed method is about \(8.1\times10^{-3}\), with an average active size of about \(4.33\) terms per output channel. FF-RLS and Param-KF use almost the full dictionary, while their prediction errors remain higher than that of the proposed method. SW-LASSO and SW-SBL give sparser models than the dense recursive estimators, but they still have larger one-step errors. Therefore, the advantage of the proposed method is not simply a lower prediction error; it is the ability to obtain that error with a compact channel-wise sparse model.

\begin{center}
\safeincludegraphics[width=0.95\linewidth,height=0.42\textheight,keepaspectratio]{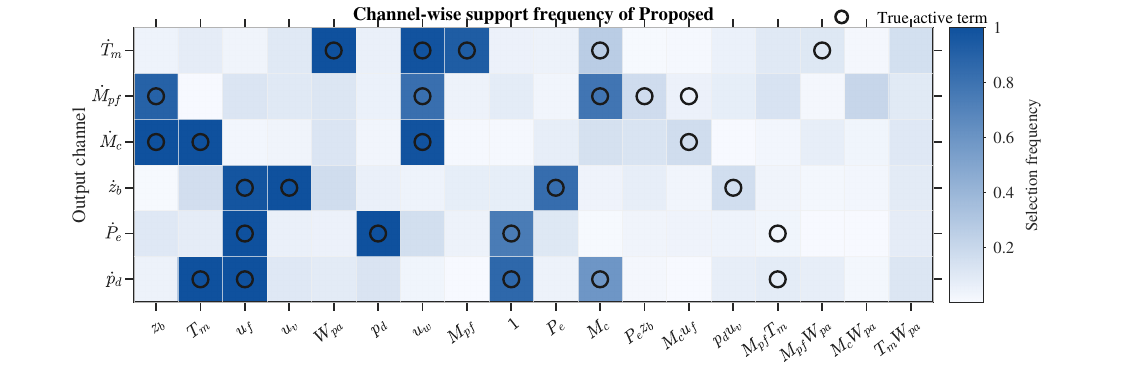}
\captionof{figure}{Channel-wise support selection frequency of the proposed method. The constant term is not shown. The color denotes the empirical selection frequency during online operation, and the circles denote the true active terms.}
\label{fig:exp2_support_frequency}
\end{center}

Fig.~\ref{fig:exp2_support_frequency} explains the sparse-model behavior at the channel level. The selected terms are concentrated near the true active dictionary terms, while most irrelevant terms have low selection frequency. This result is important because several dictionary terms are correlated through load-following operation and state--input coupling. A low one-step error alone would not be sufficient to show interpretable sparse identification. The support-frequency map shows that the proposed posterior shrinkage and active-set maintenance keep the identified model close to the intended sparse structure. Some neighboring terms are selected occasionally, which is expected in a correlated process dictionary and does not change the overall concentration of the active support.

\begin{center}
\safeincludegraphics[width=0.90\linewidth,height=0.42\textheight,keepaspectratio]{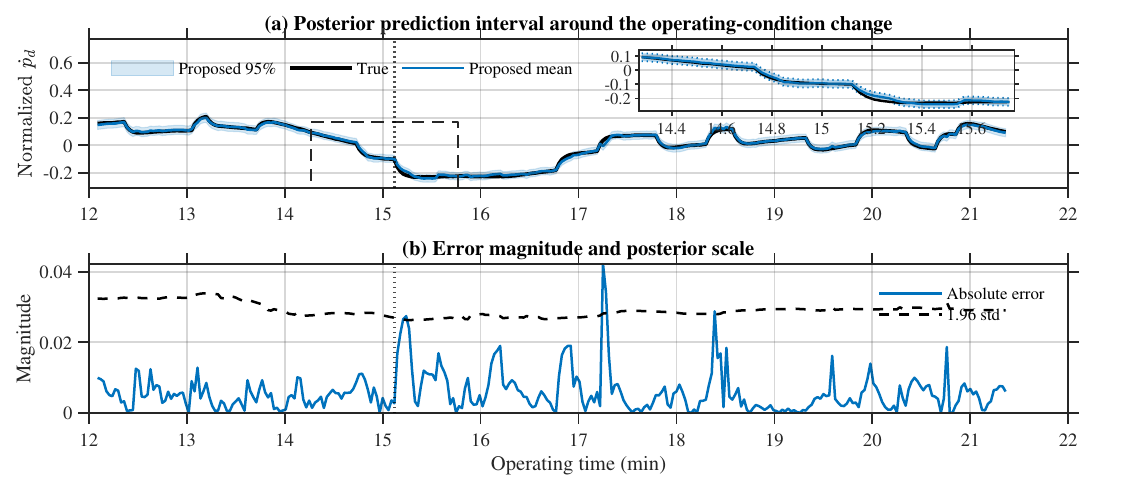}
\captionof{figure}{Channel-wise posterior prediction interval of the proposed method around the operating-change region.
(a) True value, posterior mean, and nominal $95\%$ prediction interval for the drum-pressure derivative $\dot p_d$.
(b) Absolute prediction error and posterior predictive scale.}
\label{fig:exp2_posterior_interval}
\end{center}

Fig.~\ref{fig:exp2_posterior_interval} gives a channel-wise view of the posterior uncertainty for the drum-pressure derivative $\dot p_d$. 
The posterior mean follows the true trajectory closely, and the shaded band gives the nominal $95\%$ prediction interval generated by the proposed method. 
The enlarged window in Fig.~\ref{fig:exp2_posterior_interval}(a) shows the local relation between the point prediction, the true response, and the predictive interval.

Fig.~\ref{fig:exp2_posterior_interval}(b) compares the absolute prediction error with the posterior predictive scale. 
The two quantities are of comparable magnitude over the displayed interval, which indicates that the posterior scale gives a reasonable local measure of prediction uncertainty. 
Together with Fig.~\ref{fig:exp2_dynamic_accuracy}(b), this channel-wise result illustrates how the proposed Bayesian recursion accompanies the one-step prediction with uncertainty information.

\begin{center}
\safeincludegraphics[width=0.90\linewidth,height=0.42\textheight,keepaspectratio]{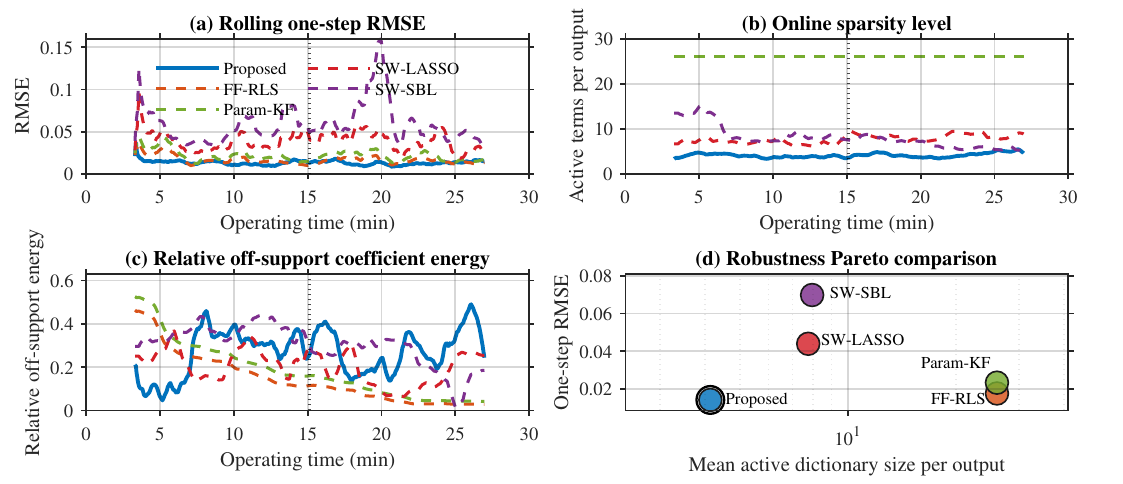}
\captionof{figure}{Additional robustness test under a noisier recursive data-generation setting. 
(a) Rolling one-step RMSE. 
(b) Online sparsity level. 
(c) Relative off-support coefficient energy. 
(d) Robustness Pareto comparison.}
\label{fig:exp2_robustness}
\end{center}

A second recursive simulation is used as an additional robustness check. It keeps the same power-plant-oriented dictionary and sparse coefficient structure, but increases the process disturbance, derivative-measurement noise, and burst probability. As shown in Fig.~\ref{fig:exp2_robustness}, the support concentration becomes less clean and the off-support energy is more affected by correlation and noise. Even under this harder setting, the proposed method keeps a compact active set and remains close to the best rolling-RMSE behavior. FF-RLS can approach the proposed method in one-step RMSE in part of this robustness test, but it does so with a dense full-dictionary model. SW-LASSO and SW-SBL remain sparser than the dense estimators but show larger prediction errors. This additional test therefore supports the main conclusion from a more demanding recursive data stream: the proposed method provides a better prediction--compactness trade-off rather than merely fitting the output with a dense adaptive model.

Overall, Experiment~2 shows that the proposed recursion can maintain a sparse nonlinear MIMO model in a power-plant-oriented dynamic setting. Compared with dense recursive estimators, it avoids unnecessary dictionary growth; compared with sliding-window sparse baselines, it gives a more favorable balance between one-step prediction and active-support compactness. The posterior covariance also responds around the operating-change region and provides a useful local indicator of prediction reliability.

\section{Conclusions}
\label{sec5}

This paper addressed the online maintenance of sparse nonlinear regressors under changing operating conditions. The proposed BRSL method updates the coefficient posterior from streaming data while preserving the sparse structure obtained after offline model construction. A sliding-window likelihood-ratio information recursion was used to incorporate new samples, remove expired samples, and discount historical information in a unified posterior update. Posterior-guided shrinkage and support maintenance were then introduced to suppress unsupported dictionary terms and allow the active structure to adapt under coefficient drift. The candidate-subspace update with an adaptive information floor ensured that the recursive posterior solve remained well posed.

The numerical results show that the proposed method maintains a favorable balance between online prediction accuracy, sparse support consistency, and posterior uncertainty. In the sparse coefficient-tracking experiment, it followed drifting coefficients while reducing off-support coefficient leakage. In the power-plant-oriented MIMO nonlinear time-varying benchmark, it produced compact and stable models under changing operating conditions, outperforming dense recursive and sliding-window sparse baselines in the main sparsity--accuracy trade-off. These results indicate that the proposed method is suitable for maintaining interpretable sparse regressors when the dictionary remains meaningful but the coefficients and active support evolve online.

\begin{ack} This work was supported by the Shanxi Province  General Program of Natural Science Research (202403021221055), Shanxi Province Major Special Program of Science and Technology (202501060301003), and Gemeng Group Technology Innovation Fund Project (2024-05, 2025-01).
\end{ack}

\printcredits

\bibliographystyle{elsarticle-num}

\bibliography{cas-refs}

\end{document}